\documentclass[11pt,a4paper,logo]{googledeepmind}

\setleftlogo[180pt]{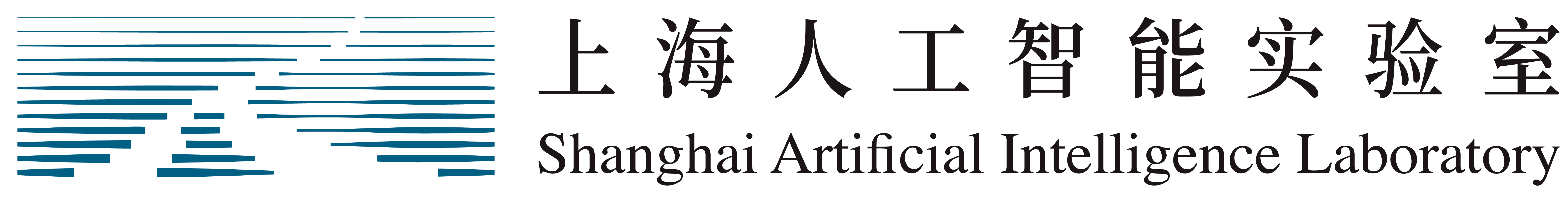} 
\setrightlogo[180pt]{}

\usepackage{placeins}
\usepackage{flafter}
\usepackage[justification=centering]{subcaption}

\usepackage[
    natbib=true,
    backend=biber,
    style=numeric, 
    sorting=none 
]{biblatex}
\AtEveryBibitem{\clearfield{month}}
\AtEveryBibitem{\clearfield{day}}

\usepackage{csquotes}
\usepackage{fontawesome5}

\providecommand{\sysname}{MARCH}
\newcommand{\safinname}{Safin-1}
\newcommand{\githuburl}{https://github.com/AI45Lab/Safin-1}
\newcommand{\huggingfaceurl}{https://huggingface.co/collections/AI45Research/safin-1}

\newcommand{\safintitlelogo}{%
    \begin{minipage}[c]{0.105\linewidth}
        \centering
        \includegraphics[width=\linewidth]{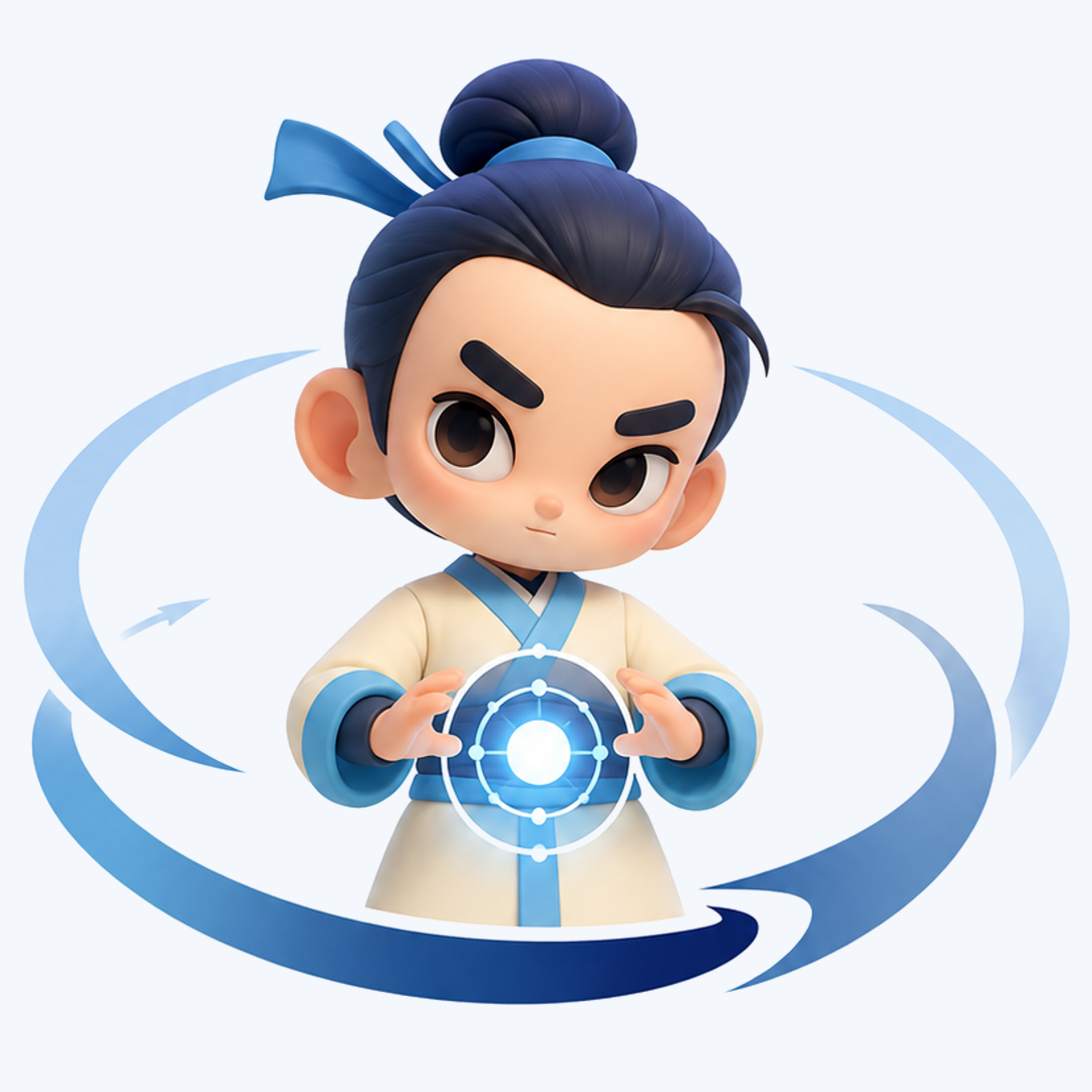}%
    \end{minipage}%
}
\title{%
    \safintitlelogo%
    \hspace{0.4em}%
    \begin{minipage}[c]{0.865\linewidth}
        \raggedright
        \fontsize{19.5}{22}\selectfont
        \safinname: Safety from Within through Memory-Native State Evolution
    \end{minipage}%
}

\usepackage{pdflscape}
\usepackage{amsthm}

\usepackage{textcomp}
\usepackage{rotating}
\usepackage{setspace}
\usepackage{soul} 
\usepackage{multicol}
\usepackage[capitalise,noabbrev]{cleveref}

\sethlcolor{green!14}

\usepackage{array}
\usepackage{multirow}
\usepackage{amsmath}
\usepackage{siunitx}

\usepackage{enumitem}
\usepackage{float}
\usepackage{seqsplit}
\usepackage{framed}
\usepackage{tikz}
\usepackage{listings}
\usepackage{colortbl}
\usepackage{wrapfig}
\usepackage[most]{tcolorbox}
\usepackage{pdfpages}
\tcbuselibrary{listingsutf8}
\definecolor{mycolor}{RGB}{50,80,150}

\definecolor{ArtifactLinkBorder}{HTML}{56C6D4}
\definecolor{ArtifactLinkBackground}{HTML}{ECFAFC}
\newtcbox{\artifactbutton}{
    on line,
    colback=ArtifactLinkBackground,
    colframe=ArtifactLinkBorder,
    boxrule=0.55pt,
    arc=3pt,
    boxsep=0pt,
    left=7pt,
    right=7pt,
    top=3pt,
    bottom=3pt,
    fontupper=\normalfont\fontsize{9}{10.5}\selectfont,
    nobeforeafter
}
\newcommand{\huggingfaceicon}{%
    \raisebox{-0.17em}{\includegraphics[height=1.15em]{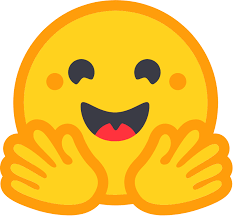}}%
}
\newcommand{\resourcebutton}[3]{%
    \artifactbutton{%
        \vphantom{\huggingfaceicon}%
        \makebox[68pt][c]{%
            \href{#1}{\textcolor{black}{#2\hspace{0.45em}#3}}%
        }%
    }%
}
\newcommand{\badgerow}{%
    \resourcebutton{\githuburl}{\faGithub}{GitHub}%
    \hspace{0.45em}%
    \resourcebutton{\huggingfaceurl}{\huggingfaceicon}{Hugging Face}%
}

\author{\fontsize{10}{12}\selectfont
\textbf{Safin Team, Shanghai AI Laboratory}%
\par\vspace{6pt}\badgerow}

\usepackage{ragged2e}
\usepackage{makecell}
\usepackage{adjustbox}

\usepackage[symbol]{footmisc}
\newcolumntype{Y}{>{\RaggedRight\arraybackslash}X}

\hypersetup{
    colorlinks=true,
    linkcolor=mycolor,
    citecolor=mycolor,
    filecolor=black,
    urlcolor=mycolor
}
\definecolor{AbstractBgColor}{HTML}{F4F7FB}
\usepackage{tocloft}

\usepackage{etoolbox}
\makeatletter
\patchcmd{\@tocline}
    {\hfil}
    {\leaders\hbox{\hfil}\hfil}
    {}{}
\makeatother

\begin{document}
\emergencystretch=1em
\clubpenalty=10000
\widowpenalty=10000
\displaywidowpenalty=10000
\thispagestyle{firststyle}

\begin{tcolorbox}[
    colback=AbstractBgColor, 
    colframe=AbstractBgColor, 
    arc=5pt,                  
    auto outer arc,
    boxrule=0pt,              
    left=8pt, right=8pt, 
    top=4pt, bottom=4pt,      
    parbox=false,
    width=\textwidth,
    before skip=-800pt,        
    after skip=0pt, 
    enlarge top by=-12pt
]
\begin{abstract}
\vspace{-10pt}
\mdseries
Long-horizon complex tasks require foundation models to accumulate
information, maintain internal states, and adapt over extended interactions.
In these settings, safety should be an intrinsic property of the model itself,
rather than a behavioral constraint that depends solely on external safeguards
or post-hoc alignment procedures such as supervised fine-tuning. This
motivates \textit{Safety from Within}, in which safety-relevant capabilities
are represented and invoked through the model's native computation rather than
relying solely on external safeguards. We present \textbf{\safinname{}}, a
family of foundation models that realizes this principle through memory
routing and state evolution. \safinname{}
is built on Memory-Anchor Routing across Context History (MARCH), a novel
network architecture that maintains structured memory states and selectively
retrieves relevant historical information through content-conditioned routing.
\safinname{} further supports test-time adaptation of
persistent capability states without repeatedly modifying the backbone model,
enabling controlled specialization over a shared general-purpose foundation.
We investigate this persistent-memory interface on downstream safety tasks
through a \textit{Safety State}, demonstrating effective state-based
adaptation with substantial safety improvements. More broadly, the routed-state
interface unifies contextual memory and persistent capability adaptation within
the model's native computation. This reframes model memory from a passive
record of prior context into an active substrate for maintaining and evolving
model behavior over time. Evaluations across
general capabilities, long-context understanding, retrieval, and efficiency
further validate \safinname{}. Together, these findings provide a concrete
path toward safety as a state-native and adaptively maintainable capability.
We emphasize that this work represents only an initial architectural exploration of \textit{Safety from Within}, and substantial further work is needed to realize this broader vision.


\end{abstract}

\newpage

\maketitle

\end{tcolorbox}

\vspace{0.6em}

\begin{figure}[H]
    \centering
    \includegraphics[width=\textwidth]{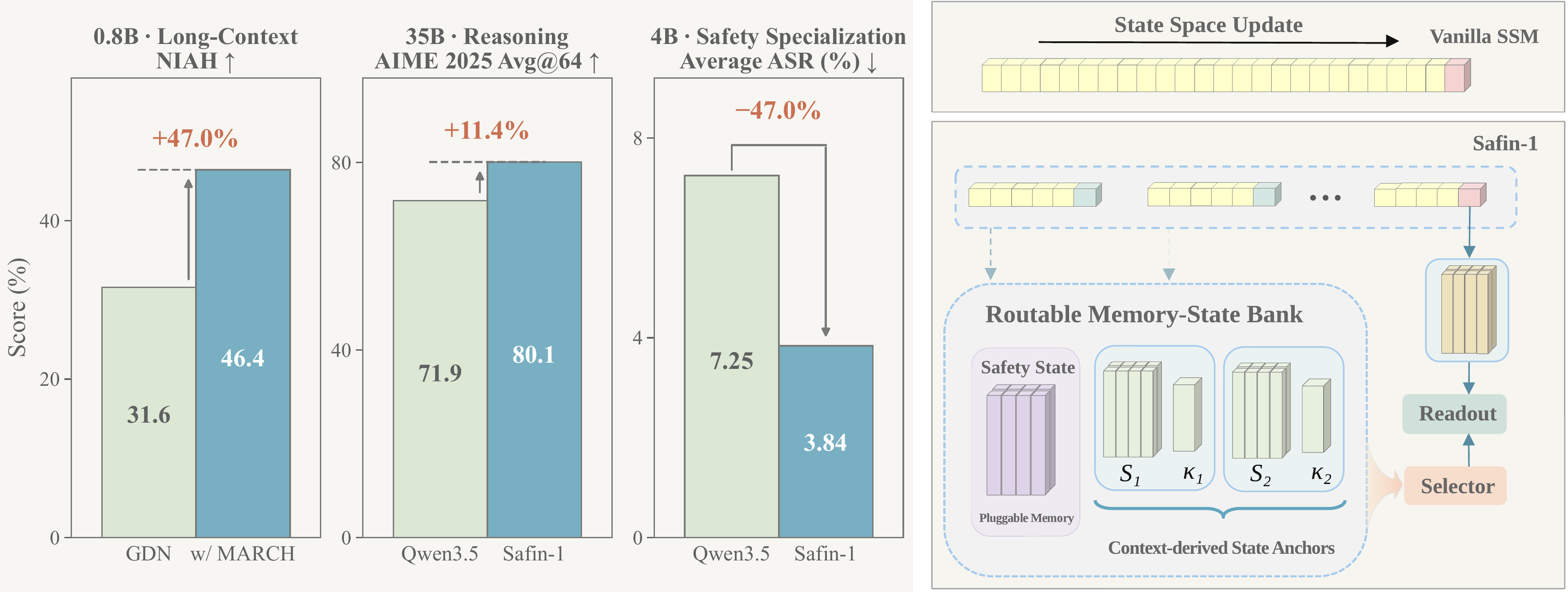}
    \caption{\safinname{} architecture and representative results. Left:
    results from 0.8B architectural validation and 35B-A3B scaling. The 4B
    safety panel compares Qwen3.5 with \safinname{} equipped with its learned
    Safety State. Right: the \safinname{} architecture routes context-derived
    and persistent memory states for selective readout.}
    \label{fig:safin-at-a-glance}
\end{figure}

\clearpage
\tableofcontents
\newpage

\setlength{\textfloatsep}{12pt plus 3pt minus 3pt}
\setlength{\floatsep}{10pt plus 2pt minus 2pt}
\setlength{\intextsep}{10pt plus 2pt minus 2pt}

\section{Introduction}
\label{sec:introd}

Long-horizon intelligence requires two forms of continuity. A model must
retain what has happened across an extended context, and it must reliably
invoke the capabilities appropriate to the current situation. Language
models typically realize these functions through different substrates:
context is retained in a token-level key--value cache or a recurrent state,
whereas reusable behaviors such as safety are commonly encoded through model
updates or adapters, or imposed at inference time through learned prefixes and
representation interventions
~\citep{bianchi2024safetytuned,li2025salora,yu2026singletoken,
wang2024inferaligner}. At a broader level, the \texttt{R$^2$AI} perspective~\citep{sun2026r2ai}
argues that scalable safety should evolve alongside model capability~\citep{yang2024towards,lab2025safework},
combining resistance to known threats with resilience to unforeseen
risks~\citep{wen2026jailbreakskill,qi2026darwin,li2026evodefense,zeng2026trace,wen2026magic}. These observations motivate a broader question: can
model state serve not only as a transient compression of context, but also as
a native substrate for persistent and selectively invoked capabilities? We
study this question through memory and safety, two settings that require a
model to recover relevant historical information while invoking appropriate
behavioral constraints without indiscriminately suppressing useful
responses~\citep{rottger2024xstest}.

Self-attention preserves fine-grained access to prior tokens but incurs
quadratic training cost and a key--value cache that grows with sequence
length~\citep{vaswani2017attention}. Recurrent and linear-attention models
instead compress the causal prefix into a fixed-size state, enabling efficient
constant-memory decoding
~\citep{katharopoulos2020transformers,dao2024transformers}. Selective
state-space models and delta-rule recurrences substantially improve how this
state is written, revised, and forgotten
~\citep{gu2023mamba,schlag2021lineartransformersselfretrievalcompressive,
yang2025gateddelta}. However, conventional recurrent designs expose only the
latest state for direct reading. Once an earlier association is attenuated or
overwritten, its previous representation may no longer be recoverable from the
current state alone~\citep{arora2024zoology,arora2024simple}. Recent work
alleviates this \emph{single-state bottleneck} by enlarging recurrent memory,
routing among multiple states, or retaining temporally organized state
collections~\citep{pan25SSE,cabannes26SDM,du2025mom,guo2026loglinear,
wang2026dynamic,behrouz26memory}. Yet retaining additional states introduces a
second challenge: historical states must be represented compactly and
retrieved according to the current context. A scalable recurrent memory
therefore requires not only state preservation, but also an efficient
state-addressing interface.

We present \textbf{\safinname{}}, a family of foundation models designed to
explore this state-native formulation of long-horizon intelligence and safety.
\safinname{} is built on \textbf{M}emory-\textbf{A}nchor \textbf{R}outing
across \textbf{C}ontext \textbf{H}istory
(\textbf{MARCH})~\citep{zhang2026march}, which preserves earlier versions of
the model's evolving recurrent state as addressable internal memory. Beyond
context-derived memories, the same routing interface can host learned
persistent capability states. We instantiate this interface as a detachable
\emph{Safety State}, allowing \safinname{} to acquire safety specialization
through test-time adaptation of model state while keeping the shared
language-model backbone frozen.

At the architectural level, the recurrent backbone periodically checkpoints
its evolving state to form addressable \emph{state anchors}. Each anchor is
paired with a compact, state-aware routing key, allowing every token to
retrieve relevant historical states or select a learned null option when
historical memory is unnecessary. Retrieved states are fused with the current
recurrent path, preserving the behavior of the underlying recurrence. This
design retains the model's own state trajectory as causal, end-to-end
trainable memory native to the recurrent computation.

This addressable state bank also provides \safinname{} with an interface for
persistent capability specialization. The bank can contain both dynamic
anchors derived from the current context and learned persistent states that
encode reusable specialization. We investigate safety as the first instance
of the latter. With the language-model backbone frozen, we learn a persistent
\emph{Safety State} and place its layer-wise states in the same bank as the
context-derived anchors; the existing router then determines their
contribution for each token. The memory can be attached or removed without
rewriting the shared backbone. Learned from harmful and benign
examples~\citep{wang2026star1}, Safety State places safety specialization
inside the model's native recurrent-state system while preserving a shared
general-purpose foundation. We view this state-native and selectively invoked
safety capability as a concrete realization of \emph{Safety from Within}, also referred to as \emph{Making Safe AI}~\citep{sun2026r2ai,dalrymple2024towards,fornasiere2026scientist,tan2025towards}.

We evaluate \safinname{} and its underlying architecture through two
complementary stages. First, controlled 0.8B studies isolate the
architectural foundation of \safinname{} by pretraining matched models on 50B
tokens and applying MARCH to Gated DeltaNet, Kimi Delta Attention, and Gated
DeltaNet-2
~\citep{yang2025gateddelta,kimi2025linear,hatamizadeh2026gateddeltanet2}.
These experiments isolate the contribution of the proposed memory-routing
architecture and establish consistent improvements in general language modeling,
LongBench~\citep{bai2024longbench}, real-world in-context retrieval, and RULER
NIAH~\citep{hsieh2024ruler}, including extrapolation beyond the training
context.

Second, we scale \safinname{} to 4B and 35B-A3B using the hybrid Qwen3.5
backbones~\citep{qwen2026qwen35} through matched continual pretraining and
supervised fine-tuning. The resulting variants raise the macro-average over
ten capability benchmarks from 66.79 to 69.20 and from 76.25 to 78.35,
respectively. The gains are most pronounced on challenging reasoning and
competition-level mathematics. Together, these results suggest that state
routing strengthens the model's
ability to sustain and recover useful computation over long reasoning horizons
without sacrificing broad general-purpose capability. The same pattern across
the dense 4B and mixture-of-experts 35B-A3B configurations further indicates
that these benefits persist across model scales and backbone organizations.

Within the scaled \safinname{} study, we additionally start from the
\safinname{} SFT checkpoints, freeze the language-model backbone, and
investigate the effectiveness of a persistent Safety State.
This state-native specialization reduces average jailbreak attack success rate by 42.3\% at 4B and 52.3\% at 35B-A3B, while producing substantially less
over-refusal than a training-matched rank-8 LoRA
control~\citep{hu2022lora}.
This study therefore represents an initial architecture-level exploration of
\emph{Safety from Within}, opening a broader research agenda for alternative
and complementary architectures that make safety native to model computation.

We additionally develop fused dense and sparse routing implementations and
evaluate their efficiency up to 128K tokens. Sparse state routing provides a
state-level analogue of sparse attention: it selects among compressed
recurrent memories rather than token-level keys, making historical retrieval
practical at long sequence lengths~\citep{lu2025moba,yuan2025native}.

Our main contributions are summarized as follows:
\begin{itemize}
  \item \textbf{\safinname{} and its state-native architecture.}
  We introduce \safinname{}, a family of foundation models built on MARCH.
  Its architecture converts the transient trajectory of recurrent computation
  into a growing bank of addressable state anchors, enabling selective access
  to earlier model states without altering the underlying recurrent update.

  \item \textbf{Persistent capability states for Safety from Within.}
  We extend the memory-state bank with learned persistent capability states
  and instantiate this interface as a detachable Safety State. This mechanism
  specializes a frozen \safinname{} backbone through its native state-routing
  pathway, making safety a model-native and selectively invoked capability.

  \item \textbf{Controlled and scaled empirical validation.}
  We validate the underlying memory-routing architecture through controlled
  0.8B pretraining studies across multiple recurrent backbones, then scale
  \safinname{} to 4B and 35B-A3B through continual pretraining and supervised
  fine-tuning. Across general capability, long-context, retrieval, safety, and
  efficiency evaluations, the results demonstrate consistent architectural
  gains and a favorable safety--over-refusal trade-off.
\end{itemize}
 
\section{Technical Methodology}
\label{sec:method}

\begin{figure}[t]
  \centering
  \includegraphics[width=\textwidth]{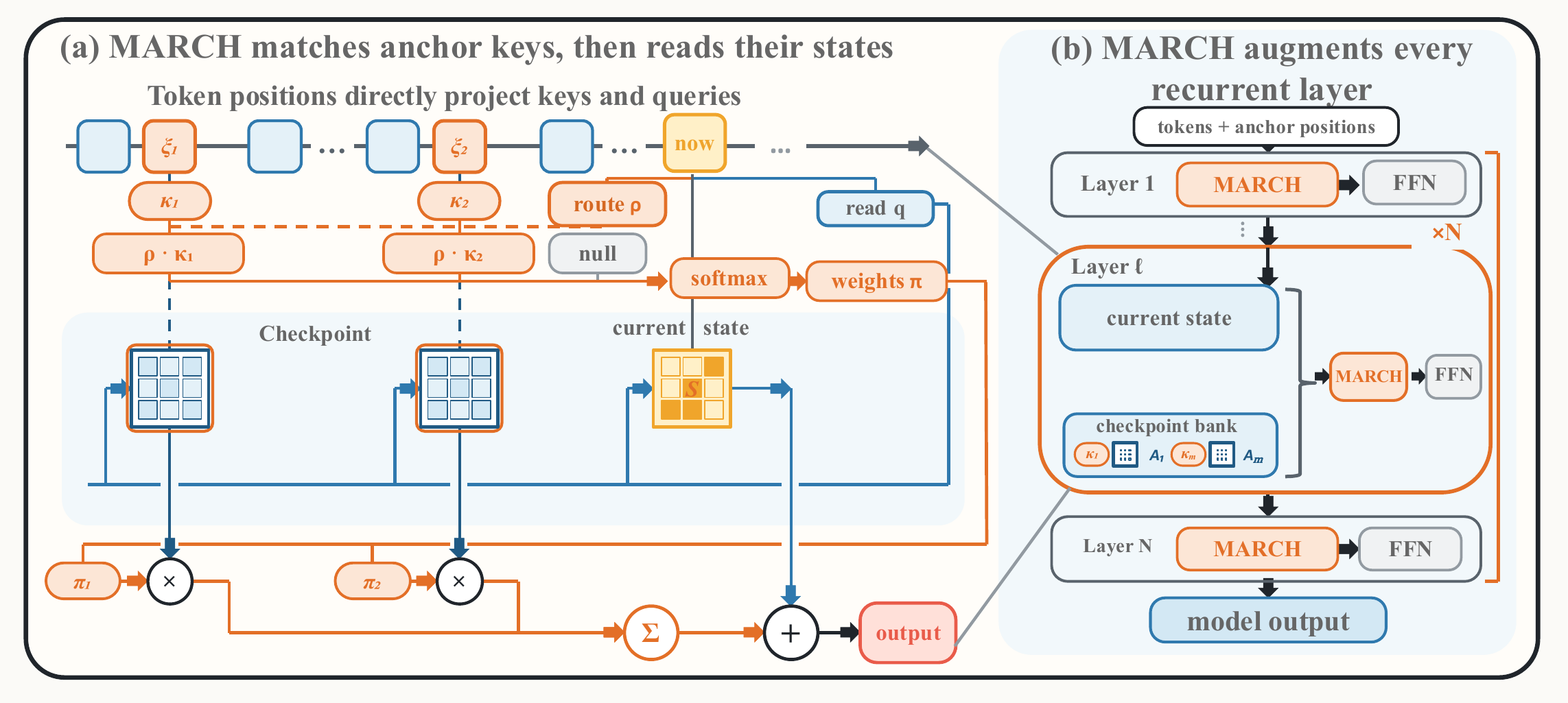}
  \caption{MARCH state-routing backbone underlying \safinname{}. Left: the
  continuously evolving recurrent state is periodically checkpointed into
  addressable state anchors. Each token routes over causally visible anchor
  keys and a learned null option, and the retrieved historical readout is
  fused with the current-state path. Right: the state-routing mechanism is
  applied at every recurrent layer, with state-aware routing keys
  reconstructed across layers.}
  \label{fig:march-architecture}
\end{figure}

\safinname{} augments recurrent computation with a unified, routable bank of
model-internal states. Its state-routing backbone is instantiated by MARCH and
comprises three components: context-derived state anchors that preserve earlier
versions of the evolving recurrent state, token-conditioned routing that
selects relevant states or a learned null option, and persistent capability
states that remain available independently of the current sequence. For each
token, the selected state readout is fused residually with the current-state
path, preserving the underlying recurrent update.
Figure~\ref{fig:march-architecture} illustrates the dynamic anchoring and
historical-reading pathway; persistent capability states are introduced in
Section~\ref{sec:method-persistent-states}.

\subsection{Context-Derived State Anchors}
\label{sec:method-anchors}
\paragraph{Anchor placement.}
Let \(\mathcal{T}=[t_1,\ldots,t_L]\) be a sequence of \(L\) text tokens.
An anchoring policy specifies an ordered set of text boundaries
\(\mathcal{B}=\{b_m\}_{m=1}^{M}\), where
\(0=b_0<b_1<\cdots<b_M\leq L\). We insert one occurrence of a shared learned
anchor embedding after each boundary:
\begin{equation}
  \widehat{\mathcal{T}}
  =
  \mathop{\mathbin{\Vert}}_{m=1}^{M}
  \Big(
    [t_{b_{m-1}+1},\ldots,t_{b_m}]
    \mathbin{\Vert}[\xi_m]
  \Big)
  \mathbin{\Vert}
  [t_{b_M+1},\ldots,t_L].
  \label{eq:march-anchors}
\end{equation}
where \(\Vert\) denotes sequence concatenation, and \(\xi_m\) is the
\(m\)-th occurrence of the shared anchor embedding \(\xi\). Text and anchor
positions have distinct computational roles. Text positions apply the base
recurrent update, allowing the state to evolve continuously across boundaries.
Immediately after processing \(t_{b_m}\), the architecture checkpoints the
resulting cumulative state. The following anchor position \(\xi_m\) does not
modify the recurrent state; instead, it reads the aligned checkpoint to
construct compact routing metadata. Each boundary therefore produces a
coupled pair: a recurrent-state snapshot and a representation through which
that snapshot can later be addressed.

\paragraph{Cumulative recurrent-state checkpointing.}
At each state-bearing layer \(\ell\), the base recurrent update remains
unchanged and carries the state
\(\mathbf{S}^{(\ell)}_t\in\mathbb{R}^{d_v\times d_k}\)
continuously across anchor boundaries. At each boundary \(b_m\), it
snapshots the current state:
\begin{equation}
  \mathbf{A}^{(m,\ell)}
  =
  \mathbf{S}^{(\ell)}_{b_m}
  \in\mathbb{R}^{d_v\times d_k},
  \qquad m=1,\ldots,M.
  \label{eq:march-anchor}
\end{equation}
Because the recurrence is not reset at anchor boundaries,
\(\mathbf{A}^{(m,\ell)}\) represents the cumulative prefix up to position
\(b_m\), rather than only the most recent segment. We therefore refer to it as
a \emph{state anchor}. The ordered bank
\(\{\mathbf{A}^{(1,\ell)},\ldots,\mathbf{A}^{(M,\ell)}\}\) preserves the
temporal trajectory of a single recurrent memory before later updates
attenuate or overwrite earlier contents.

\paragraph{Content-conditioned anchor metadata.}
Let $\mathbf{u}^{(\ell)}_m$ denote the normalized input representation of
anchor position $\xi_m$ at layer $\ell$. The anchor position reads only
its aligned state checkpoint:
\begin{equation}
  \mathbf{q}^{(\ell)}_m
  =
  \mathbf{W}^{(\ell)}_q\mathbf{u}^{(\ell)}_m,
  \qquad
  \mathbf{o}^{(\ell)}_m
  =
  \mathbf{A}^{(m,\ell)}\mathbf{q}^{(\ell)}_m.
  \label{eq:march-anchor-token-read}
\end{equation}
The same input representation is projected into a compact routing key:
\begin{equation}
  \boldsymbol{\kappa}^{(\ell)}_m
  =
  \mathbf{W}^{(\ell)}_k
  \mathbf{u}^{(\ell)}_m
  \in\mathbb{R}^{d_r}.
  \label{eq:march-routing-key}
\end{equation}
The aligned readout is passed through the standard output projection and
residual pathway, making the next-layer anchor representation
$\mathbf{u}^{(\ell+1)}_m$ dependent on its associated checkpoint
$\mathbf{A}^{(m,\ell)}$. Consequently, although all anchor positions share the
same input embedding, their representations---and therefore their routing
keys---become state-dependent after the first layer. The router thus addresses
anchors according to their retained contents rather than their temporal
indices alone.

\subsection{Content-Routed State Retrieval}
\label{sec:method-routing}

\paragraph{Candidate scoring and null routing.}
For a text token at position $t$, the causally available state anchors are
indexed by $\mathcal{V}_t=\{m\in\{1,\ldots,M\}\mid b_m<t\}$. In the
dynamic-memory setting, these anchors form the non-null candidate set
$\mathcal{C}_t=\mathcal{V}_t$. The routing module projects the normalized
hidden state $\mathbf{x}_t$ into a routing query and scores it against the key
of each candidate:
\begin{equation}
  \boldsymbol{\rho}_t
  =
  \mathbf{W}_R\mathbf{x}_t,
  \qquad
  a_{t,j}
  =
  \boldsymbol{\rho}_t^{\top}\boldsymbol{\kappa}_j,
  \quad j\in\mathcal{C}_t.
  \label{eq:march-routing-score}
\end{equation}
To allow the model to bypass state retrieval, we augment the candidate set
with a null option $\varnothing$, whose payload is fixed to
zero, $\mathbf{A}^{(\varnothing)}=\mathbf{0}$. Its query-dependent
logit is $n_t=\mathbf{w}_{\varnothing}^{\top}\mathbf{x}_t
+b_{\varnothing}$.
Let $\widetilde{\mathcal{C}}_t
=\mathcal{C}_t\cup\{\varnothing\}$ denote the augmented candidate set.
We define the logit of each candidate $j\in\widetilde{\mathcal{C}}_t$ as
\begin{equation}
  s_{t,j}
  =
  \begin{cases}
    a_{t,j}, & j\in\mathcal{C}_t,\\
    n_t,     & j=\varnothing,
  \end{cases}
  \qquad
  \pi_{t,j}
  =
  \frac{\exp(s_{t,j})}
       {\displaystyle
        \sum_{r\in\widetilde{\mathcal{C}}_t}
        \exp(s_{t,r})}.
  \label{eq:march-routing-prob}
\end{equation}
The routing query $\boldsymbol{\rho}_t$ determines which candidates to
retrieve, whereas the state-read query $\mathbf{q}_t$ reads their
matrix-valued contents. Because the null candidate has a zero payload and
participates in the same softmax, probability assigned to it attenuates the
retrieval branch without introducing a separate fusion gate. If
$\mathcal{C}_t$ is empty, the null candidate receives all probability mass.

\paragraph{Retrieved-state fusion.}
Given the routing probabilities, the candidate-state readouts are weighted
and added to the current-state readout using the underlying recurrence's
native state-read query $\mathbf{q}_t$:
\begin{equation}
  \mathbf{o}_t
  =
  \mathbf{S}_t\mathbf{q}_t
  +
  \sum_{j\in\widetilde{\mathcal{C}}_t}
  \pi_{t,j}
  \mathbf{A}^{(j)}\mathbf{q}_t.
  \label{eq:march-core}
\end{equation}
This additive formulation preserves the original recurrent path and
introduces state retrieval as an auxiliary branch without modifying the
underlying recurrent update. Because the routing probabilities directly
affect the layer output, the routing queries and state-derived keys are
optimized end-to-end with the language-modeling objective.

We note that the aggregation formulation in Equation
\eqref{eq:march-routing-prob} readily admits a sparse variant by restricting
aggregation to the $K$ highest-scoring visible anchors (Top-$K$). This sparse
approach exhibits natural connections with hierarchical sparse attention
approaches \citep{lu2025moba, HiLS}, while preserving dense token-level
processing rather than relying on hard token-level pruning. Our ablation
studies in Section~\ref{sec:march-ablations} show that sparse routing
substantially reduces aggregation cost with minimal performance degradation.

\subsection{Persistent Capability States}
\label{sec:method-persistent-states}

Beyond context-derived anchors, the routed state bank at each state-bearing
layer can also host $J$ learnable persistent capability states,
$\{\mathbf{P}_{j}\}_{j=1}^{J}$. We suppress the layer index in this subsection
for clarity. Each persistent state has the same matrix-valued structure as a
dynamic anchor, and its routing key $\boldsymbol{\kappa}_{p_j}$ is generated
through the same state-aware metadata pathway defined in
Equations~\eqref{eq:march-anchor-token-read}--\eqref{eq:march-routing-key}.
Unlike dynamic anchors, persistent states are learned parameters rather than
checkpoints constructed from the current sequence. They are therefore
available from the first text token without consuming positions in the input
sequence.

We extend the non-null candidate set as
\begin{equation}
  \mathcal{C}_t
  =
  \mathcal{V}_t \cup \{p_1,\ldots,p_J\},
  \qquad
  \mathbf{A}^{(p_j)}
  =
  \mathbf{P}_{j},
  \label{eq:march-persistent-candidates}
\end{equation}
where $p_j$ identifies the $j$-th persistent candidate. Dynamic anchors and
persistent states are scored by the same routing query and jointly normalized
with the null option through a single softmax.
Consequently, Equations~\eqref{eq:march-routing-prob} and
\eqref{eq:march-core} operate directly on the extended candidate set. The
router therefore assigns each persistent state a token- and context-dependent
contribution rather than adding it with a fixed weight at every position.

In \safinname{}, we instantiate this interface for safety. The layer-wise
collection $\{\mathbf{P}_{\mathrm{safe}}^{(\ell)}\}_{\ell}$ constitutes the
\emph{Safety State}. During specialization, we keep the language-model
backbone frozen and optimize only this collection; the training data and
optimization setup are detailed in Section~\ref{sec:safety-state}. Once
learned, the Safety State can be attached to or removed from the routed state
bank without changing the shared backbone parameters. This state-routed
specialization provides the mechanism through which \safinname{} realizes
\emph{Safety from Within}.

\subsection{Efficient Producer--Reader Implementation}
\label{sec:method-efficient-training}
The state-routing computation is organized into a recurrent producer and a
fused state reader. The producer inherits the hardware-efficient chunkwise
kernel of the underlying recurrent mixer. In the GDN-based implementation
studied here, it follows the chunkwise formulation of Gated
DeltaNet~\citep{yang2025gateddelta}, processing recurrent updates in
tensor-core-friendly blocks while producing the current-state readout and
checkpointing the recurrent state at anchor boundaries. These checkpoints,
together with any persistent capability states, form the candidates consumed
by the state reader.

Following the I/O-aware principles of
FlashAttention~\citep{dao2022flashattention}, the reader jointly tiles query
tokens and state candidates, reuses each candidate tile across a block of
queries, and fuses routing-score computation, normalization, and weighted
state readout into a streaming reduction. This schedule avoids materializing
either the dense token-to-candidate routing matrix or the substantially larger
tensor of per-candidate matrix-valued readouts, reducing intermediate storage
and HBM traffic.

The same producer supports both dense and Top-$K$ retrieval. The sparse path
leaves the recurrent computation and compact routing-score calculation
unchanged, while restricting the more expensive matrix-valued state readout
to the routed candidates. \safinname{} uses Top-$4$ routing by default. As
shown in Section~\ref{sec:efficiency}, at 128K tokens, Top-$4$ routing more
than doubles end-to-end throughput relative to dense routing and reduces core
runtime by roughly an order of magnitude.
 
\providecommand{\sysname}{MARCH}

\section{Small-Scale Validation of MARCH}
\label{sec:march-validation}

We first isolate the architectural contribution of MARCH through controlled
0.8B pretraining studies across multiple recurrent backbones. By matching the
training data, token budget, sequence length, and optimization protocol, these
experiments test whether addressable historical states improve language
modeling, long-context retrieval, and length extrapolation independently of
model scale. The corresponding training configuration is provided in
Appendix~\ref{app:small-scale-training}.

\subsection{Experimental Setup}

\paragraph{Model Suite and Training Protocol.}
Our recurrent baselines include Gated DeltaNet (GDN)~\citep{yang2025gateddelta},
GDN augmented with Log-Linear Attention~\citep{guo2026loglinear}, Kimi Delta
Attention (KDA)~\citep{kimi2025linear}, and Gated DeltaNet-2
(GDN2)~\citep{hatamizadeh2026gateddeltanet2}. We further apply \sysname{} to
GDN, KDA, and GDN2 to evaluate its compatibility across different recurrent
update rules. All recurrent models consist of 21 layers, with a hidden size
of 1,536 and six value heads, totaling approximately 0.8B parameters. We also
include two full-attention baselines for comparison: a depth-matched 21-layer
model with 693M parameters and an approximately parameter-matched 24-layer
model with 778M parameters. Both use 16 attention heads and a RoPE base of
500K. Following the academic-scale protocol of Log-Linear Attention, all
models are pretrained from scratch on the same 50B-token subset of
Long-Data-Collections, with a maximum sequence length of 16K. For \sysname{},
we set the routing dimension to $d_r=64$ and place state anchors every
$C=512$ tokens. All other training configurations are held fixed across
sequence mixers and summarized in Table~\ref{tab:small-scale-training-config}.

\paragraph{Evaluation.}
We evaluate four complementary capability groups. \emph{General language modeling} is assessed on eight zero-shot commonsense benchmarks: LAMBADA
\citep{paperno_lambada_2016}, PIQA~\citep{bisk_piqa_2020}, HellaSwag
\citep{zellers_hellaswag_2019}, WinoGrande
\citep{sakaguchi_winogrande_2021}, ARC-Easy and ARC-Challenge
\citep{clark_think_2018}, OpenBookQA~\citep{OpenBookQA2018}, and
CommonsenseQA~\citep{talmor2019commonsenseqaquestionansweringchallenge}.
\emph{Long-context understanding} is evaluated on
LongBench~\citep{bai2024longbench}. For \emph{controlled long-context retrieval
and length extrapolation}, we evaluate three single-needle and three
multi-needle NIAH tasks from RULER~\citep{hsieh2024ruler} at context lengths of
4K, 8K, 16K, 32K, and 64K. Since all models are pretrained with a maximum
sequence length of 16K, performance at 32K and 64K measures zero-shot length
extrapolation. Finally, \emph{in-context retrieval} is evaluated on six
real-world, recall-intensive datasets: SQuAD
\citep{rajpurkar_know_2018}, TriviaQA~\citep{JoshiTriviaQA2017},
SWDE~\citep{lockard_openceres_2019}, FDA~\citep{arora_language_2023}, Natural
Questions~\citep{kwiatkowski-etal-2019-natural}, and
DROP~\citep{dua2019drop}. We follow the evaluation protocol of prior
work~\citep{wang2026dynamic} and use LM-Evaluation-Harness
\citep{eval-harness} where applicable.

\subsection{Language Modeling and Long-Context Capabilities}

\subsubsection{General Language Modeling}

\begin{table}[!t]
  \centering
  \caption{Zero-shot performance of MARCH and baseline models on eight
commonsense reasoning benchmarks. Results are reported using accuracy
(\texttt{acc}) or normalized accuracy (\texttt{acc\_n}), as indicated in
the column headers; higher is better
(\textcolor{ForestGreen}{$\uparrow$}). The best result among Gated DeltaNet variants in each column is highlighted in bold.}
  \label{tab:lm_cs}
  \small
  \begin{tabular}{l|cccccccc|c}
      \toprule
      \textbf{Model}
      & \textbf{LMB.} & \textbf{PIQA} & \textbf{Hella.} & \textbf{Wino.}
      & \textbf{ARC-e} & \textbf{ARC-c} & \textbf{OBQA} & \textbf{CSQA}
      & \textbf{Avg.} \\
      & acc \textcolor{ForestGreen}{$\uparrow$}
      & acc \textcolor{ForestGreen}{$\uparrow$}
      & acc \textcolor{ForestGreen}{$\uparrow$}
      & acc \textcolor{ForestGreen}{$\uparrow$}
      & acc\_n \textcolor{ForestGreen}{$\uparrow$}
      & acc\_n \textcolor{ForestGreen}{$\uparrow$}
      & acc\_n \textcolor{ForestGreen}{$\uparrow$}
      & acc \textcolor{ForestGreen}{$\uparrow$}
      & \\
      \midrule
      Transformer
      & 49.4 & 66.5 & 33.9 & 52.1 & 47.8
      & 26.4 & 32.0 & 22.1 & 41.3 \\
      \quad w/ \emph{24 Layers}
      & 50.3 & 67.6 & 34.4 & 50.6 & 46.3
      & 25.8 & 31.2 & 24.7 & 41.4 \\
      \midrule
      Gated DeltaNet
      & 48.5 & 66.1 & 33.1 & 50.8 & 45.9 & 25.3
      & 30.0 & 21.1 & 40.1 \\
      \quad w/ \emph{Log-Linear}
      & 47.7 & 65.7 & 33.2 & 51.9 & 44.3 & 24.9
      & 30.4 & 21.7 & 40.0 \\
      \quad w/ \emph{\sysname}
      & \textbf{49.5} & \textbf{66.9} & \textbf{34.8} & \textbf{52.6}
      & \textbf{47.1} & \textbf{25.6} & \textbf{32.8} & \textbf{22.5}
      & \textbf{41.5} \\
      \bottomrule
  \end{tabular}
\end{table}

Table~\ref{tab:lm_cs} reports results on all eight zero-shot benchmarks. Among the recurrent models, \sysname{} achieves the best performance on every benchmark, raising the average score to 41.5, compared with 40.1 for Gated DeltaNet and 40.0 for its Log-Linear variant. Notably, \sysname{} also slightly outperforms both full-attention reference models on average. These results indicate that incorporating historical-state retrieval preserves general language modeling capability while improving access to long-range information. The improvement is broad rather than driven by a single outlier: \sysname{} exceeds the strongest recurrent baseline on each of the eight tasks, spanning both completion-oriented and discriminative commonsense evaluations. This consistency supports interpreting the aggregate gain as preservation of base capability rather than benchmark-specific specialization.

\FloatBarrier
\subsubsection{Length Extrapolation and Associative Recall}
\begin{figure}[H]
\centering
\includegraphics[width=\textwidth]{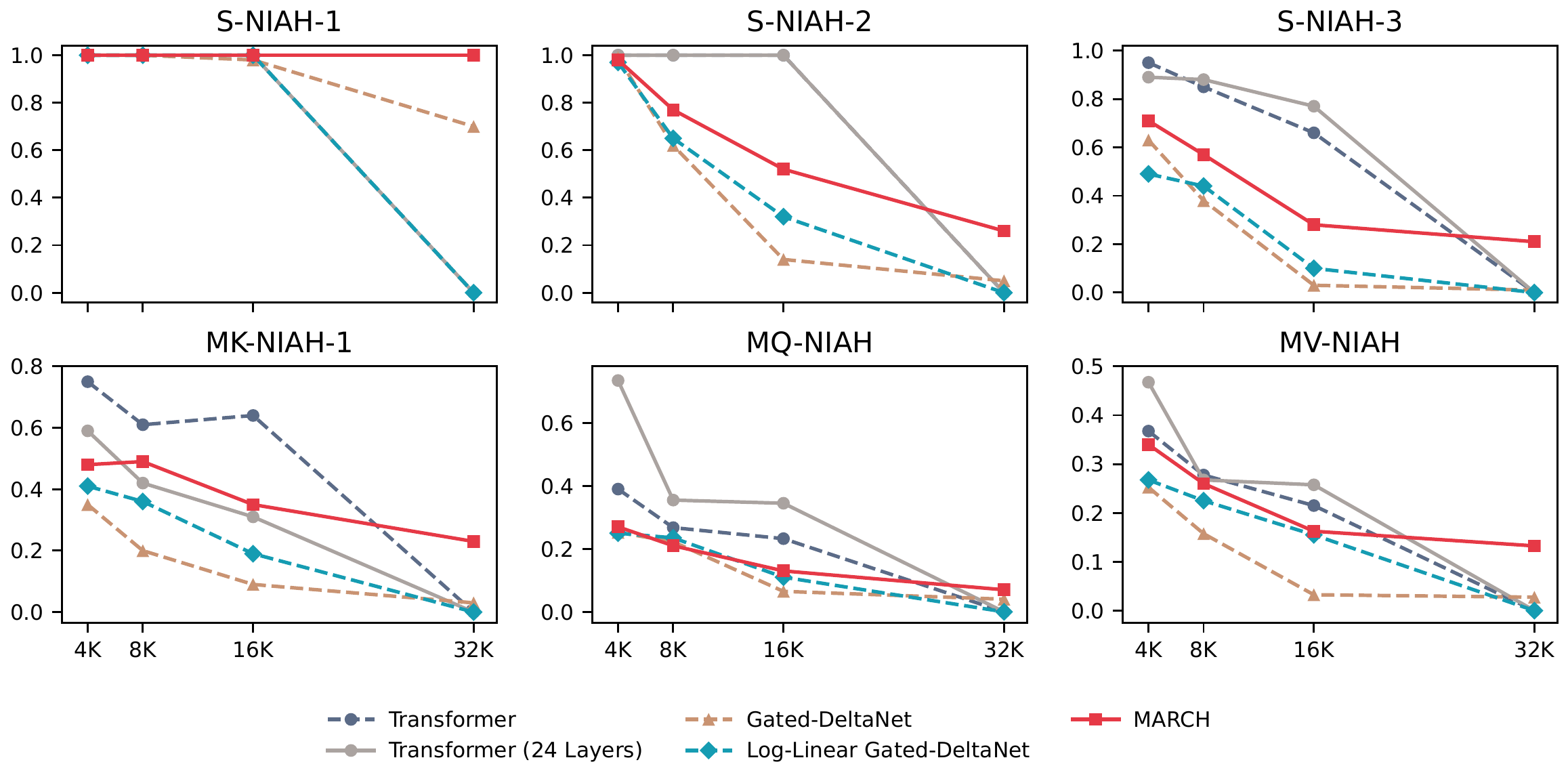}
\caption{NIAH performance on three single-needle and three multi-needle tasks. The Transformer
achieves perfect accuracy on both S-NIAH-1 and S-NIAH-2 at context lengths of 4K, 8K, and 16K.}
\label{fig:niah-lineplot}
\end{figure}

We further evaluate associative recall using the NIAH suite from
RULER~\citep{hsieh2024ruler}. Across the 24 task--length combinations in
Figure~\ref{fig:niah-lineplot}, \sysname{} outperforms the strongest recurrent
baseline in 19 settings and matches it in the remaining five. The gains are
most pronounced on multi-needle tasks and at 32K, beyond the maximum training
length of 16K. This behavior is consistent with the routing construction:
because the same content router is shared across all state anchors, extending
the sequence introduces additional retrievable states without requiring
anchor-specific or length-specific parameters. The 32K results therefore
suggest that the retrieval mechanism generalizes beyond the training context
length, a property not captured by short-context performance alone.

\FloatBarrier

\subsubsection{Long-Context Understanding and In-Context Retrieval}

\begin{table}[!t]
\centering
\caption{Results on twelve LongBench tasks. The
best result among Gated DeltaNet and its variants is shown in bold for each
task. The relative average gain over the Gated DeltaNet is shown in
green parentheses.}
\label{tab:long_bench}
\small
\setlength{\tabcolsep}{3.2pt}
\renewcommand{\arraystretch}{1.08}
\begin{tabular}{@{}l|ccc|ccc|ccc|ccc|c@{}}
\toprule
& \multicolumn{3}{c|}{\textbf{Single-Doc QA}} & \multicolumn{3}{c|}{\textbf{Multi-Doc QA}} & \multicolumn{3}{c|}{\textbf{Summarization}} & \multicolumn{3}{c|}{\textbf{Few-shot Learning}} & \\ \midrule
 \textbf{Model} & {NQA} & {QQA} & {MFQ} 
          & {HQA} & {2WM} & {Mus} 
          & {GvR} & {QMS} & {MNs}
          & {TRC} & {TQA} & {SSM}
          & \textbf{Avg.\textcolor{ForestGreen}{$\uparrow$}} \\ \midrule
Transformer               & 4.4 & 4.1 & 15.9 & 7.5 & 9.9 & 4.1 & 10.7 & 11.6 & 14.5 & 21.0 & 33.9 & 28.3 & 13.8 \\
\quad w/ \emph{24 Layers} & 3.5 & 11.1 & 18.3 & 7.8 & 9.8 & 4.1 & 11.3 & 12.9 & 12.8 & 22.5 & 47.5 & 23.2 & 15.4 \\
\midrule
Gated DeltaNet             & 3.0 & 4.8 & 13.3 & 5.6 & 8.7 & 2.1 & 2.6 & 11.3 & 13.0 & 18.0 & 38.6 & 21.9 & 11.9 \\
\quad w/ \emph{Log-Linear} & 3.6 & 6.2 & 13.6 & 7.1 & 8.2 & 3.3 & 6.3 & 13.2 & 13.5 & 17.0 & 32.8 & 25.1 & 12.5 \\
\quad w/ \emph{\sysname}  &
\textbf{4.2} &
\textbf{7.8} &
\textbf{14.6} &
\textbf{7.4} &
\textbf{11.5} &
\textbf{4.8} &
\textbf{8.2} &
\textbf{17.4} &
\textbf{14.1} &
\textbf{19.0} &
\textbf{43.1} &
\textbf{26.3} &
\textbf{14.9 (\textcolor{ForestGreen}{$\uparrow25\%$})}\\
\bottomrule
\end{tabular}
\end{table}

Table~\ref{tab:long_bench} presents results across all twelve LongBench tasks,
covering single-document QA, multi-document QA, summarization, and few-shot
learning.

\FloatBarrier

\sysname{} is the strongest recurrent model on every task, improving
the overall average to 14.9, compared with 11.9 for Gated DeltaNet and 12.5
for Log-Linear Attention. The largest category-level gains occur in
multi-document QA and summarization, where successful prediction requires
integrating information distributed across distant portions of the context.
\begin{table}[H]
\centering
\caption{In-context retrieval accuracy
(\textcolor{ForestGreen}{$\uparrow$}). The best result among Gated DeltaNet and
its variants is marked in bold for each benchmark, with the relative gain over
the stronger of the two Gated DeltaNet baselines shown in green parentheses.}
\label{tab:in-context-retrieval}
\footnotesize
\setlength{\tabcolsep}{2pt}
\renewcommand{\arraystretch}{1.05}
\begin{tabular}{@{}l|cccccc|c@{}}
\toprule
\textbf{Model}
& \textbf{SQuAD\textcolor{ForestGreen}{$\uparrow$}}
& \textbf{SWDE\textcolor{ForestGreen}{$\uparrow$}}
& \textbf{FDA\textcolor{ForestGreen}{$\uparrow$}}
& \textbf{TriviaQA\textcolor{ForestGreen}{$\uparrow$}}
& \textbf{DROP\textcolor{ForestGreen}{$\uparrow$}}
& \textbf{NQ\textcolor{ForestGreen}{$\uparrow$}}
& \textbf{Avg.\textcolor{ForestGreen}{$\uparrow$}} \\
\midrule
Transformer & 41.3 & 59.3 & 80.4 & 2.2 & 2.9 & 1.7 & 31.3 \\
\quad w/ \emph{24 Layers} & 40.4 & 64.9 & 83.7 & 4.0 & 3.4 & 2.5 & 33.2 \\
\midrule
Gated DeltaNet & 34.8 & 45.0 & 31.4 & 1.1 & 2.2 & 0.8 & 19.2 \\
\quad w/ \emph{Log-Linear} & 33.7 & 46.1 & 38.2 & 1.3 & 2.6 & 1.0 & 20.5 \\
\quad w/ \emph{\sysname}
& \textbf{37.7 (\textcolor{ForestGreen}{$\uparrow8\%$})}
& \textbf{51.9 (\textcolor{ForestGreen}{$\uparrow13\%$})}
& \textbf{44.6 (\textcolor{ForestGreen}{$\uparrow17\%$})}
& \textbf{1.6 (\textcolor{ForestGreen}{$\uparrow23\%$})}
& \textbf{2.9 (\textcolor{ForestGreen}{$\uparrow12\%$})}
& \textbf{1.2 (\textcolor{ForestGreen}{$\uparrow20\%$})}
& \textbf{23.3 (\textcolor{ForestGreen}{$\uparrow14\%$})} \\
\bottomrule
\end{tabular}
\end{table}

Table~\ref{tab:in-context-retrieval} presents results on the six real-world
in-context retrieval datasets. \sysname{} outperforms both recurrent baselines
on every dataset, raising the best baseline average from 20.5 to 23.3.
Together with the controlled RULER results, these improvements indicate that
routed state anchors benefit retrieval beyond synthetic associative-recall
settings, extending to heterogeneous real-world contexts.

\subsection{Architecture Analysis}
\label{sec:march-ablations}

\paragraph{Generality across recurrent parameterizations.}
We further evaluate \sysname{} on GDN, KDA, and GDN2, which progressively
increase the granularity and flexibility of recurrent memory updates:
KDA refines GDN with channel-wise decay, while GDN2 further decouples
erase and write operations. This progression allows us to examine whether
the benefit of \sysname{} persists as the underlying recurrent update
becomes more expressive. As shown in Table~\ref{tab:cross-backbone-comparison}, \sysname{} improves
NIAH in 11 of the 12 backbone--length settings. It also consistently improves SWDE and FDA, while largely preserving
short-context performance and training perplexity. These results indicate that \sysname{} remains effective across recurrent
updates with increasingly fine-grained memory control.

\begin{table}[!t]
\centering
\caption{
Comparison of GDN, KDA, and GDN2 with and without \sysname{}.
NIAH scores are averaged over six needle-in-a-haystack tasks.
The best result in each column is shown in bold.
}
\label{tab:cross-backbone-comparison}

\small
\setlength{\tabcolsep}{4.2pt}
\renewcommand{\arraystretch}{1.08}

\begin{tabular}{@{}l
                c
                cc
                ccc
                ccccc@{}}
\toprule
& \textbf{Train}
& \multicolumn{2}{c}{\textbf{Short-Context}\textcolor{ForestGreen}{$\uparrow$}}
& \multicolumn{3}{c}{\textbf{In-Context Retrieval}\textcolor{ForestGreen}{$\uparrow$}}
& \multicolumn{5}{c}{\textbf{NIAH (6-Task Avg.)}\textcolor{ForestGreen}{$\uparrow$}} \\
\cmidrule(lr){2-2}
\cmidrule(lr){3-4}
\cmidrule(lr){5-7}
\cmidrule(l){8-12}

\textbf{Model}
& \textbf{PPL}\textcolor{ForestGreen}{$\downarrow$}
& \textbf{LAMB.}
& \textbf{Hella.}
& \textbf{SQuAD}
& \textbf{SWDE}
& \textbf{FDA}
& \textbf{4K}
& \textbf{8K}
& \textbf{16K}
& \textbf{32K}
& \textbf{Avg.} \\
\midrule

GDN
& 6.796
& 48.52 & 33.10
& 34.85 & 45.00 & 31.40
& 53.80 & 35.67 & 21.73 & 15.13
& 31.58 \\

\quad w/ \emph{\sysname{}}
& 6.751
& 49.52 & 34.84
& 37.70 & 51.85 & 44.56
& 59.58 & 54.96 & 39.46 & \textbf{31.71}
& \textbf{46.43} \\

\addlinespace[3pt]

KDA
& 6.691
& 52.11 & 35.41
& 38.64 & 51.94 & 37.93
& 51.17 & 34.75 & 22.54 & 18.08
& 31.64 \\

\quad w/ \emph{\sysname{}}
& 6.703
& \textbf{52.57} & 34.95
& 33.58 & \textbf{54.10} & 42.83
& 58.96 & 52.25 & 32.13 & 22.17
& 41.38 \\

\addlinespace[3pt]

GDN2
& 6.663
& 51.00 & \textbf{35.58}
& \textbf{39.11} & 51.76 & 38.20
& \textbf{61.83} & 46.88 & 34.38 & 19.00
& 40.52 \\

\quad w/ \emph{\sysname{}}
& \textbf{6.610}
& 51.41 & 35.36
& 38.71 & 53.65 & \textbf{48.46}
& 57.58 & \textbf{56.00} & \textbf{40.46} & 25.17
& 44.80 \\

\bottomrule
\end{tabular}
\end{table}

\paragraph{Positional encoding and length extrapolation.}
We study RoPE in hybrid models with a 3:1 recurrent-to-attention layer ratio, trained with a context length of 16K. Evaluations at 32K and 64K therefore directly measure length extrapolation. Table~\ref{tab:shortconv-rope-ablation} shows that removing RoPE improves extrapolation for both \sysname{} and GDN and avoids the sharp degradation beyond the training context.\pagebreak[4] For GDN, however, this improvement comes at the cost of lower retrieval performance within 4K--16K. In contrast, \sysname{} benefits from removing RoPE across all context lengths, yielding both strong in-distribution retrieval and substantially smoother extrapolation to 32K and 64K.

\FloatBarrier

\begin{table}[!htbp]
  \centering
  \caption{
Effect of RoPE in hybrid models trained up to 16K context. Results at 32K and 64K evaluate length extrapolation.
The Transformer baseline is a parameter-matched 24-layer model. NIAH Avg. is averaged across the five context lengths.
The best result in each column is shown in bold.
  }
  \label{tab:shortconv-rope-ablation}

  \small
  \setlength{\tabcolsep}{3.0pt}
  \renewcommand{\arraystretch}{1.08}

  \begin{tabular}{@{}lc
                  ccc
                  cccccc@{}}
    \toprule
    & &
    \multicolumn{3}{c}{\textbf{In-Context Retrieval}\textcolor{ForestGreen}{$\uparrow$}} &
    \multicolumn{6}{c}{\textbf{NIAH (6-Task Avg.)}\textcolor{ForestGreen}{$\uparrow$}} \\
    \cmidrule(lr){3-5}
    \cmidrule(l){6-11}

    \textbf{Model}
    & \textbf{RoPE}
    & \textbf{SQuAD}
    & \textbf{SWDE}
    & \textbf{FDA}
    & \textbf{4K}
    & \textbf{8K}
    & \textbf{16K}
    & \textbf{32K}
    & \textbf{64K}
    & \textbf{Avg.} \\
    \midrule

    \multirow{2}{*}{\textbf{\sysname{}}}
      & $\checkmark$
      & 40.82 & 58.42 & 78.58
      & 76.88 & 72.29 & 66.88
      & 13.79 & 4.13 & 46.79 \\

      & $\times$
      & 41.12 & \textbf{65.44} & 80.76
      & \textbf{87.79} & \textbf{79.92} & \textbf{74.54}
      & \textbf{68.79} & \textbf{30.25} & \textbf{68.26} \\

    \addlinespace[3pt]

    \multirow{2}{*}{\textbf{GDN}}
      & $\checkmark$
      & \textbf{43.40} & 58.42 & 76.77
      & 79.96 & 73.58 & 65.79
      & 13.29 & 9.08 & 48.34 \\

      & $\times$
      & 40.11 & 65.17 & 75.23
      & 68.00 & 61.04 & 55.63
      & 48.46 & 21.33 & 50.89 \\

    \midrule

    \textbf{Transformer}
      & $\checkmark$
      & 40.48 & 64.99 & \textbf{83.76}
      & 78.04 & 65.38 & 66.88
      & 0.00 & 0.00 & 42.06 \\

    \bottomrule
  \end{tabular}
\end{table}

\paragraph{Effect of the anchor interval.}
The anchor interval $C$ controls the temporal resolution of the state bank:
shorter intervals yield denser historical anchors, but increase storage and
routing overhead. We vary $C$ from 256 to 2,048 while keeping all other model
and training settings fixed. Table~\ref{tab:chunk-size-ablation} reports
performance on three in-context retrieval benchmarks and the average over six
NIAH tasks at 4K, 8K, and 16K context lengths. We further evaluate whether a
model trained with $C=512$ can operate at different state-bank resolutions at
inference time, including the Fenwick-tree organization adopted by Log-Linear
Attention~\citep{guo2026loglinear,Fenwick1994AND}.

When training and inference resolutions are matched, $C=512$ provides the best overall trade-off. Increasing $C$ reduces the temporal resolution of the state bank and generally degrades retrieval, whereas further decreasing it to $C=256$ yields little consistent benefit despite introducing more anchors. Notably, the $C=512$ checkpoint can use a denser state bank at inference time, with $C=256$ achieving the strongest aggregate performance. Performance deteriorates as the bank becomes increasingly sparse, while the Fenwick-tree variant remains competitive despite using a state organization unseen during training, indicating that the learned historical-state reader transfers across different state-bank layouts.

These results also expose a practical deployment control: the same checkpoint
can trade retrieval quality against state-storage and routing cost by changing
only the inference-time bank construction.

\pagebreak[4]
\noindent\begin{minipage}{\linewidth}
  \centering
  \captionsetup{skip=4pt}
  \captionof{table}{Anchor-interval ablation. Panel (a) matches training and
  inference; panel (b) varies inference for the $C=512$ checkpoint. NIAH
  averages six tasks per length; column-wise best results are bold.}
  \label{tab:chunk-size-ablation}
  \small
  \setlength{\tabcolsep}{7pt}
  \begin{tabular}{@{}cc ccc ccc@{}}
    \toprule
    \multirow{2}{*}{\makecell{\textbf{Anchor}\\\textbf{Interval }$C$}}
    & \multirow{2}{*}{\makecell{\textbf{\# Anchors}\\\textbf{at 16K}}}
    & \multicolumn{3}{c}{\textbf{In-Context Retrieval}\,\textcolor{ForestGreen}{$\uparrow$}}
    & \multicolumn{3}{c}{\textbf{NIAH (6-Task Avg.)}\,\textcolor{ForestGreen}{$\uparrow$}} \\
    \cmidrule(lr){3-5}
    \cmidrule(lr){6-8}
    &
    & \textbf{SQuAD}
    & \textbf{SWDE}
    & \textbf{FDA}
    & \textbf{4K}
    & \textbf{8K}
    & \textbf{16K} \\
    \midrule
    \multicolumn{8}{l}{\textbf{(a) Matched training and inference anchor intervals}} \\
    \addlinespace[1pt]
    256
    & 64 & 36.76 & 49.23 & \textbf{44.83}
    & 58.17 & 49.25 & \textbf{44.83} \\
    512
    & 32 & \textbf{37.70} & \textbf{51.85} & 44.56
    & \textbf{59.58} & \textbf{54.96} & 39.46 \\
    1024
    & 16 & 36.49 & 48.51 & 43.28
    & 53.21 & 43.21 & 33.54 \\
    2048
    & 8 & 34.15 & 44.19 & 30.76
    & 59.54 & 39.83 & 27.96 \\
    \midrule
    \multicolumn{8}{l}{\textbf{(b) Inference-time anchor interval}} \\
    \addlinespace[1pt]
    64
    & 256 & 39.41 & 53.11 & 46.01
    & \textbf{63.33} & 55.00 & 39.67 \\
    128
    & 128 & 39.08 & 52.84 & 46.91
    & \textbf{63.33} & \textbf{57.33} & 39.17 \\
    256
    & 64 & \textbf{40.35} & \textbf{53.38} & \textbf{47.46}
    & 62.83 & 56.17 & \textbf{40.83} \\
    512
    & 32 & 37.70 & 51.85 & 44.56
    & 59.58 & 54.96 & 39.46 \\
    1024
    & 16 & 37.23 & 43.74 & 34.57
    & 55.33 & 48.33 & 32.67 \\
    2048
    & 8 & 37.23 & 33.93 & 23.23
    & 51.00 & 40.33 & 34.00 \\
    Fenwick Tree
    & 6 & 36.16 & 36.79 & 22.35
    & 52.17 & 39.56 & 33.14 \\
    \bottomrule
  \end{tabular}
\end{minipage}

\paragraph{Routing design.}
We ablate three components of the routing mechanism: the query--key dimension $d_r$, routing sparsity, and the learned null option. Table~\ref{tab:routing-ablation} reports macro-averaged results on general language understanding, LongBench, and in-context retrieval, together with NIAH performance. Our default configuration uses dense routing with $d_r=64$ and includes the null option.

\begin{table}[!htbp]
  \centering
  \caption{Routing-design ablations. The first row is the default;
  each subsequent row changes one component. Commonsense (CS), LongBench,
  and Retrieval are macro-averages over 8, 12, and 6 benchmarks,
  respectively. NIAH scores are averaged over six tasks, and Avg. over the
  three context lengths. Column-wise best results are bold.}
  \label{tab:routing-ablation}
  \small
  \setlength{\tabcolsep}{3.5pt}
  \begin{tabular}{@{}lcc ccc cccc@{}}
    \toprule
    \multicolumn{3}{c}{\textbf{Configuration}}
    & \multicolumn{3}{c}{\textbf{Benchmark Averages}\,\textcolor{ForestGreen}{$\uparrow$}}
    & \multicolumn{4}{c}{\textbf{NIAH (6-Task Avg.)}\,\textcolor{ForestGreen}{$\uparrow$}} \\
    \cmidrule(lr){1-3}
    \cmidrule(lr){4-6}
    \cmidrule(lr){7-10}
    \textbf{Routing}
    & \textbf{$d_r$}
    & \textbf{Null}
    & \textbf{CS}
    & \textbf{LongBench}
    & \textbf{Retrieval}
    & \textbf{4K}
    & \textbf{8K}
    & \textbf{16K}
    & \textbf{Avg.} \\
    \midrule
    Dense (default) & 64 & Yes
    & \textbf{41.48} & \textbf{14.87} & 23.31
    & \textbf{59.58} & \textbf{54.96} & \textbf{39.46} & \textbf{51.33} \\
    \midrule
    Dense & 192 & Yes
    & 40.94 & 13.88 & \textbf{24.52}
    & 56.71 & 51.33 & 37.38 & 48.47 \\
    Top-$4$ & 64 & Yes
    & 41.38 & 13.79 & 23.17
    & 57.29 & 46.04 & 31.21 & 44.85 \\
    Dense & 64 & No
    & 41.04 & 14.11 & 22.86
    & 54.88 & 47.92 & 35.13 & 45.98 \\
    \bottomrule
  \end{tabular}
\end{table}

Increasing $d_r$ from 64 to 192 improves in-context retrieval but degrades the other aggregate metrics, making $d_r=64$ a better overall trade-off. Top-$4$ routing largely preserves general language understanding and retrieval performance, but underperforms dense routing on NIAH, suggesting a viable efficiency--accuracy trade-off rather than a stronger default. Finally, removing the null option consistently degrades performance across all aggregates, indicating that explicitly allowing the router to ignore irrelevant historical states is important for robust routing.

\FloatBarrier

\subsection{Efficiency and Scalability}
\label{sec:efficiency}

\paragraph{Training efficiency.}
Figure~\ref{fig:training-efficiency} compares end-to-end throughput and core
forward--backward runtime for FlashAttention-2, Gated DeltaNet, dense
\sysname{}, and Top-$4$ \sysname{}. Sparse routing becomes increasingly
beneficial as the context grows. At 128K tokens, Top-$4$ \sysname{} more than
doubles the training throughput of dense \sysname{} and reduces its core
runtime by roughly an order of magnitude. It also exceeds FlashAttention-2 in
throughput at this length, although vanilla Gated DeltaNet remains faster
because it does not perform historical retrieval.

\begin{center}
  \begin{minipage}{\textwidth}
    \centering
    \includegraphics[width=\linewidth]{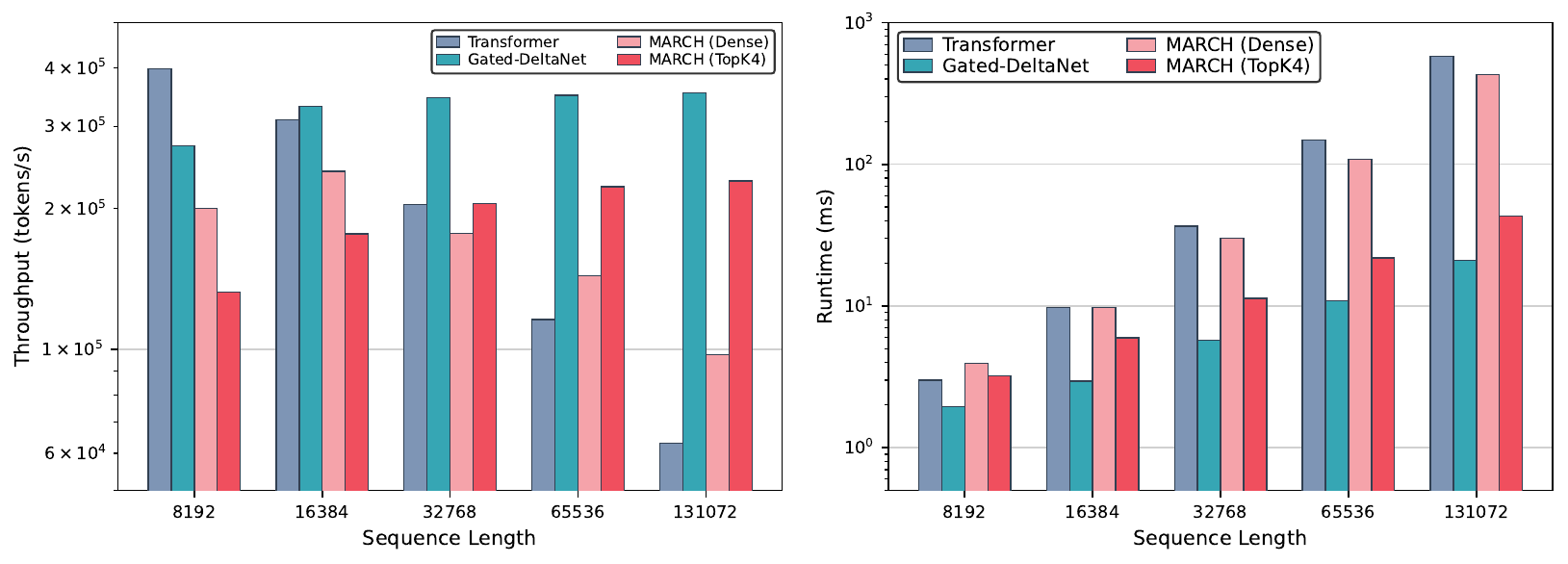}
    \captionsetup{hypcap=false}
    \captionof{figure}{Training efficiency across sequence lengths. Left:
    end-to-end training throughput in tokens per second (higher is better).
    Right: forward--backward runtime of the core sequence-mixing operation in
    milliseconds (lower is better). \sysname{} (Top-$4$) retains only the four
    highest-scoring state anchors for each token and head during historical
    retrieval.}
    \label{fig:training-efficiency}
  \end{minipage}
\end{center}

Taken together, these controlled studies validate the effectiveness,
generality, and scalability of MARCH's dynamic state routing, demonstrating
consistent gains in language modeling, retrieval, and length extrapolation
while retaining practical efficiency at long sequence lengths.

\FloatBarrier

\section{\safinname{} at Scale}
\label{sec:safin-at-scale}

We next examine whether the gains observed under controlled pretraining carry
over to larger foundation models, and whether \safinname{} can support
persistent capability specialization through its routed state bank. We
instantiate \safinname{} at the 4B and 35B-A3B scales, first evaluating the
resulting models after matched continual pretraining and supervised
fine-tuning, and then studying Safety from Within by adapting a persistent
Safety State while keeping the shared backbone frozen.

\subsection{Continual Pretraining and Supervised Fine-Tuning}
\label{sec:safin-scaling}

\subsubsection{Experimental Setup}

\paragraph{Models and training protocol.}

We instantiate \safinname{} at two scales from the dense Qwen3.5-4B and
Qwen3.5-35B-A3B mixture-of-experts checkpoints~\citep{qwen2026qwen35}. Both
backbones employ a hybrid architecture that interleaves Gated DeltaNet (GDN)
and full-attention layers. Architecturally, \safinname{} equips 16 selected
GDN layers with the MARCH state-routing mechanism, comprising state anchoring
and content-routed historical readout, while retaining the underlying GDN
recurrence, full-attention layers, and overall hybrid layout. Historical
retrieval uses Top-$4$ routing to control routing cost as the state bank grows.

Within each scale, the Qwen3.5 baseline and its \safinname{} counterpart are
initialized from the same checkpoint and trained with matched data mixtures,
token budgets, sequence lengths, and optimization settings. We first perform
continual pretraining for 50B tokens on the Intern-S2 pre-training
mixture~\citep{bai2026interns2preview}, followed by supervised fine-tuning
for a 30B-token budget using
\href{https://huggingface.co/datasets/nvidia/Nemotron-Cascade-2-SFT-Data}{Nemotron-Cascade-2-SFT-Data}~\citep{yang2026nemotroncascade2}.
The SFT mixture spans general instruction following, knowledge and reasoning,
mathematics, code, and safety alignment. Both stages use a maximum sequence
length of 32K. Unless otherwise noted, we report results from the final SFT
checkpoints. Complete model and training configurations are provided in
Tables~\ref{tab:large-scale-model-config} and
\ref{tab:large-scale-training-config} in
Appendix~\ref{app:safin-training}.

\paragraph{Evaluation benchmarks.}
We evaluate the final SFT checkpoints across five evaluation categories:
\begin{itemize}[leftmargin=1.4em,itemsep=0.1em,topsep=0.2em]
  \item \textbf{Knowledge and reasoning:}
  MMLU~\citep{hendrycks2021mmlu},
  MMLU-Pro~\citep{wang2024mmlupro}, and
  GPQA-Diamond~\citep{rein2024gpqa}.
  \item \textbf{Mathematics:}
  GSM8K~\citep{cobbe2021gsm8k},
  MATH~\citep{hendrycks2021math}, and AIME 2025.
  \item \textbf{Instruction following and long-context understanding:}
  IFEval~\citep{zhou2023ifeval} and
  LongBench v2~\citep{bai2025longbench}.
  \item \textbf{Code generation:}
  MBPP~\citep{austin2021mbpp} and
  HumanEval~\citep{chen2021humaneval}.
  \item \textbf{Safety robustness and over-refusal:}
  We report attack success rate (ASR; lower is better) on
  WildJailbreak~\citep{jiang2024wildteaming},
  FORTRESS~\citep{knight2025fortress},
  StrongREJECT~\citep{souly2024strongreject},
  Jailbreak-R1~\citep{guo2025jailbreakr1}, and
  JailbreakBench (JBB)~\citep{chao2024jailbreakbench}. We additionally report
  over-refusal rate (ORR; lower is better) on benign prompts from
  XSTest~\citep{rottger2024xstest}.
\end{itemize}
These safety benchmarks evaluate \safinname{} and Qwen3.5 following the shared
CPT--SFT pipeline. Section~\ref{sec:safety-state} separately studies the
effect of augmenting \safinname{} with a persistent Safety State.

\subsubsection{Capability and Safety Evaluation}
Table~\ref{tab:qwen-sft-results} compares \safinname{} with its matched Qwen3.5
counterparts across capability and safety benchmarks at both model scales.

\begin{table}[!t]
  \centering
  \caption{Capability and safety results after the matched CPT--SFT pipeline
  at the 4B and 35B-A3B scales. All values are percentages. Bold denotes the
  better result within each scale; ties are bolded for both models. ASR and
  ORR are lower-is-better, whereas all other metrics are higher-is-better.}
  \label{tab:qwen-sft-results}
  \small
  \setlength{\tabcolsep}{9pt}
  \renewcommand{\arraystretch}{1.08}
  \begin{tabular}{@{}lcccc@{}}
    \toprule
    \multirow{2}{*}{\textbf{Task}}
      & \multicolumn{2}{c}{\textbf{4B}}
      & \multicolumn{2}{c}{\textbf{35B-A3B}} \\
    \cmidrule(lr){2-3}\cmidrule(l){4-5}
      & \textbf{Qwen3.5} & \textbf{\safinname}
      & \textbf{Qwen3.5} & \textbf{\safinname} \\
    \midrule
    \rowcolor{gray!12}[0pt][0pt]
    \multicolumn{5}{@{}l@{}}{\textbf{Knowledge and Reasoning}} \\
    MMLU
      & \textbf{75.47} & 74.55 & 81.14 & \textbf{81.83} \\
    MMLU-Pro
      & 58.27 & \textbf{67.55} & 75.33 & \textbf{75.98} \\
    GPQA-Diamond
      & 57.58 & \textbf{58.59} & 64.65 & \textbf{71.21} \\
    \midrule
    \rowcolor{gray!12}[0pt][0pt]
    \multicolumn{5}{@{}l@{}}{\textbf{Mathematics}} \\
    GSM8K
      & \textbf{87.72} & 87.11 & 87.87 & \textbf{89.31} \\
    MATH
      & 89.22 & \textbf{89.90} & \textbf{94.10} & 93.58 \\
    AIME 2025 (Avg@64)
      & 55.57 & \textbf{63.59} & 71.88 & \textbf{80.10} \\
    \midrule
    \rowcolor{gray!12}[0pt][0pt]
    \multicolumn{5}{@{}l@{}}{\textbf{Instruction Following and Long-Context Understanding}} \\
    IFEval
      & 76.52 & \textbf{76.71} & 78.37 & \textbf{79.67} \\
    LongBench v2
      & 35.19 & \textbf{35.39} & 42.54 & \textbf{42.94} \\
    \midrule
    \rowcolor{gray!12}[0pt][0pt]
    \multicolumn{5}{@{}l@{}}{\textbf{Code}} \\
    MBPP
      & 59.20 & \textbf{65.40} & 77.60 & \textbf{78.60} \\
    HumanEval
      & \textbf{73.17} & \textbf{73.17} & 89.02 & \textbf{90.24} \\
    \midrule
    \rowcolor{gray!12}[0pt][0pt]
    \multicolumn{5}{@{}l@{}}{\textbf{Safety Robustness}} \\
    WildJailbreak
      & 4.40 & \textbf{3.60} & \textbf{5.20} & 6.80 \\
    FORTRESS
      & 24.80 & \textbf{23.20} & 23.80 & \textbf{17.80} \\
    StrongREJECT
      & \textbf{1.28} & \textbf{1.28} & 1.28 & \textbf{0.64} \\
    Jailbreak-R1
      & 4.79 & \textbf{3.19} & 4.15 & \textbf{3.19} \\
    JBB
      & \textbf{1.00} & 2.00 & \textbf{0.00} & 1.00 \\
    \textbf{Average ASR}
      & 7.25 & \textbf{6.65} & 6.89 & \textbf{5.89} \\
    \midrule
    \rowcolor{gray!12}[0pt][0pt]
    \multicolumn{5}{@{}l@{}}{\textbf{Over-Refusal (ORR $\downarrow$)}} \\
    XSTest
      & 8.80 & \textbf{8.60} & \textbf{16.80} & 17.60 \\
    \bottomrule
  \end{tabular}
\end{table}

We highlight five main findings.

\textit{(i) \safinname{} improves challenging reasoning benchmarks while
preserving broad knowledge performance.}
At the 4B scale, \safinname{} raises MMLU-Pro from 58.27 to 67.55
(+9.28 points) and GPQA-Diamond from 57.58 to 58.59 (+1.01 points), with
only a 0.92-point decrease on MMLU. At the 35B-A3B scale, \safinname{} improves
all three benchmarks, most notably raising GPQA-Diamond by 6.56 points
(from 64.65 to 71.21). These results show that \safinname{} preserves broad
knowledge performance while delivering larger gains on more challenging
reasoning tasks.

\textit{(ii) \safinname{} improves competition-level mathematical reasoning
while maintaining performance on established mathematics benchmarks.}
On AIME 2025, \safinname{} raises Avg@64 by 8.02 points at 4B and 8.22 points
at 35B-A3B. Meanwhile, its performance on GSM8K and MATH remains stable: the
absolute difference from the matched Qwen3.5 baseline is at most 0.68 points
at 4B and 1.44 points at 35B-A3B. These results indicate that the substantial
AIME gains do not come at the expense of broader mathematical capability.

\textit{(iii) \safinname{} yields consistent gains in instruction following
and long-context understanding.}
\safinname{} improves both IFEval and LongBench v2 at the two model scales.
At 4B, the gains are 0.19 and 0.20 points, respectively; at 35B-A3B, they
increase to 1.30 and 0.40 points. Although modest, the consistent improvements
across both benchmarks show that \safinname{} preserves these capabilities
across both the dense 4B and larger MoE configurations.

\textit{(iv) \safinname{} maintains or improves code generation across model
scales.}
At 4B, \safinname{} raises MBPP by 6.20 points while matching the Qwen3.5
baseline on HumanEval. At 35B-A3B, it improves MBPP and HumanEval by 1.00 and
1.22 points, respectively. These results show that \safinname{} preserves
code-generation capability and provides additional gains across both model
scales.

Across these four capability groups, the pattern remains consistent across the
dense 4B and MoE 35B-A3B backbones. Gains concentrate on harder reasoning and
structured-generation tasks, while established-benchmark performance remains
broadly stable.

\Needspace{7\baselineskip}
\textit{(v) \safinname{} improves average safety robustness with limited
change in over-refusal.}
Across the five jailbreak benchmarks, average ASR falls by 0.60 points at 4B
and 1.00 point at 35B-A3B, with the largest gain on FORTRESS at 35B-A3B
(23.80 to 17.80). Despite small regressions on WildJailbreak and JBB at this
scale, aggregate jailbreak resistance improves. XSTest ORR changes from 8.80
to 8.60 at 4B and from 16.80 to 17.60 at 35B-A3B.

\paragraph{Overall.}
\safinname{} improves seven of the ten capability benchmarks at 4B, with one
tie, and nine of the ten at 35B-A3B. Together with the average
reductions in jailbreak ASR, these results show that \safinname{} delivers
broad capability gains and improved average safety robustness after continual
pretraining and supervised fine-tuning, without a systematic trade-off in
general capability or safety.

\FloatBarrier
\subsection{Safety from Within via a Persistent Safety State}
\label{sec:safety-state}

Following Section~\ref{sec:method-persistent-states}, we instantiate a
persistent Safety State across the state-bearing \safinname{} layers and
evaluate whether this state-native specialization improves safety while
preserving the capabilities of the frozen backbone.

\subsubsection{Training Data and Experimental Setup}

\paragraph{Training data and context construction.}
We construct a safety-specialization corpus from 1,000 harmful requests paired
with safety-aligned responses in STAR-1 and 915 benign requests paired with
helpful responses in STAR-benign~\citep{wang2026star1}. Together, these sources
provide 1,915 distinct target conversations. For each target, we create three
context variants by prepending 0, 512, or 2,048 tokens of benign context,
producing 5,745 training examples in total. Each non-empty prefix is assembled
exclusively from complete benign user--assistant turns.

Beyond standard safety alignment on isolated prompts, this construction
explicitly trains the Safety State under different historical-memory
configurations. Because \safinname{} checkpoints the cumulative recurrent state
every 512 tokens to form a state anchor, the 0-, 512-, and 2,048-token
conditions present the same target with no context-derived state anchor, one
state anchor, and a bank of multiple state anchors, respectively. Reusing the
same target supervision across these conditions controls for target content
while varying the amount and configuration of historical memory. This
encourages the learned Safety State to remain effective across different
context lengths and state-anchor bank compositions. Finally, joint supervision
on harmful and benign targets couples safety with helpfulness: the model must
produce safety-aligned responses to harmful requests while continuing to
answer legitimate requests, thereby reducing the tendency toward
indiscriminate refusal.

\paragraph{Persistent Safety State training.}
We initialize the 4B and 35B-A3B experiments from the corresponding
\safinname{} SFT checkpoints and freeze all language-model parameters. The only
trainable component is a persistent Safety State, represented by a persistent
state inserted into the routable memory-state bank of each state-bearing
\safinname{} layer. We optimize the persistent states with standard
token-level causal cross-entropy.

\paragraph{Training-matched LoRA control.}
For a controlled comparison, we train a rank-8 LoRA control~\citep{hu2022lora}
from the same SFT checkpoint using the same data, optimizer, batch size, and
number of updates. The Safety State lane is disabled for this control, while periodic state
anchoring and content-routed historical retrieval remain active.
We use $r=8$, $\alpha=32$, and dropout 0.05 without bias training or
RSLoRA. At 4B, adapters cover the recurrent, full-attention, and MLP
projections; at 35B-A3B, they cover the recurrent and full-attention
projections but no dense, expert, or shared-expert MLP projections.

\paragraph{Evaluation protocol.}
We reuse the safety suite introduced in
Section~\ref{sec:safin-scaling}. The
five jailbreak benchmarks are evaluated by attack success rate (ASR), while
XSTest measures the over-refusal rate (ORR) on benign prompts; lower is better for
both. To distinguish genuine safety improvement from indiscriminate refusal or
broader capability loss, we additionally report retention on MMLU-Pro, MATH
(Minerva), AIME 2025 (Avg@64), and MBPP. Optimizer settings and numerical precision are summarized in
Table~\ref{tab:safety-state-training-config} in
Appendix~\ref{app:safety-state-training}.

\subsubsection{Safety--Utility Trade-off}

We compare the persistent Safety State with the unadapted \safinname{}
checkpoint and the training-matched LoRA control at both model scales. The
comparison covers jailbreak robustness, over-refusal on benign prompts, and
retention of general capabilities. Table~\ref{tab:safety-state-results}
summarizes the results, from which three findings emerge.

\begin{table}[H]
  \centering
  \caption{Safety--utility trade-off of the persistent Safety State and
  training-matched rank-8 LoRA control for \safinname{} at the 4B and
  35B-A3B scales.}
  \label{tab:safety-state-results}
  \small
  \renewcommand{\arraystretch}{1.04}
  \setlength{\tabcolsep}{3pt}

  \begin{tabularx}{\linewidth}{@{}
    >{\raggedright\arraybackslash}p{0.22\linewidth}
    >{\hsize=0.78\hsize\centering\arraybackslash}X
    >{\hsize=1.44\hsize\centering\arraybackslash}X
    >{\hsize=0.78\hsize\centering\arraybackslash}X
    >{\hsize=0.78\hsize\centering\arraybackslash}X
    >{\hsize=1.44\hsize\centering\arraybackslash}X
    >{\hsize=0.78\hsize\centering\arraybackslash}X@{}}
    \toprule
    \multirow{2}{*}{\textbf{Task}}
      & \multicolumn{3}{c}{\textbf{\safinname{} (4B)}}
      & \multicolumn{3}{c}{\textbf{\safinname{} (35B-A3B)}} \\
    \cmidrule(lr){2-4}\cmidrule(l){5-7}
      & \textbf{Base}
      & \mbox{\textbf{Safety State}}
      & \textbf{LoRA}
      & \textbf{Base}
      & \mbox{\textbf{Safety State}}
      & \textbf{LoRA} \\
    \midrule

    \rowcolor{gray!12}[0pt][0pt]
    \multicolumn{7}{@{}l@{}}{\textbf{Safety Robustness (ASR \textcolor{ForestGreen}{$\downarrow$})}} \\
    WildJailbreak
      & 3.60 & 2.40 & \textbf{1.60}
      & 6.80 & 1.60 & \textbf{1.20} \\
    FORTRESS
      & 23.20 & \textbf{13.60} & 20.20
      & 17.80 & \textbf{11.80} & 12.40 \\
    StrongREJECT
      & 1.28 & \textbf{1.60} & 2.88
      & 0.64 & \textbf{0.32} & 0.64 \\
    JailBreak-R1
      & 3.19 & 1.60 & \textbf{0.96}
      & 3.19 & \textbf{0.32} & 0.96 \\
    JBB
      & 2.00 & \textbf{0.00} & 1.00
      & 1.00 & \textbf{0.00} & \textbf{0.00} \\
    \addlinespace[1pt]
    \textbf{Average ASR}
      & 6.65 & \textbf{3.84} & 5.33
      & 5.89 & \textbf{2.81} & 3.04 \\

    \addlinespace[2pt]
    \rowcolor{gray!12}[0pt][0pt]
    \multicolumn{7}{@{}l@{}}{\textbf{Over-refusal (ORR \textcolor{ForestGreen}{$\downarrow$})}} \\
    XSTest
      & 8.60 & \textbf{9.00} & 13.60
      & 17.60 & \textbf{19.60} & 27.60 \\

    \addlinespace[2pt]
    \rowcolor{gray!12}[0pt][0pt]
    \multicolumn{7}{@{}l@{}}{\textbf{Capability Retention (\textcolor{ForestGreen}{$\uparrow$})}} \\
    MMLU-Pro
      & 67.55 & \textbf{66.49} & 65.55
      & 75.98 & 72.61 & \textbf{74.92} \\
    MATH
      & 89.90 & \textbf{87.54} & 87.42
      & 93.58 & \textbf{93.32} & 93.12 \\
    AIME 2025 (Avg@64)
      & 63.59 & \textbf{60.99} & 60.42
      & 80.10 & \textbf{74.21} & 72.61 \\
    MBPP
      & 65.40 & \textbf{64.40} & 62.80
      & 78.60 & 74.40 & \textbf{76.80} \\
    \addlinespace[1pt]
    \textbf{Average}
      & 71.61 & \textbf{69.86} & 69.05
      & 82.07 & 78.64 & \textbf{79.36} \\
    \bottomrule
  \end{tabularx}

  \begin{minipage}{\linewidth}
    \footnotesize
    \textit{Note.} Base is the unadapted \safinname{} checkpoint. Averages are
    unweighted means across the five safety or four capability benchmarks,
    respectively. Bold compares Safety State with LoRA within each scale; ties
    are bolded for both. All values are percentages.
  \end{minipage}
\end{table}

\textit{(i) The persistent Safety State achieves the lowest average jailbreak
ASR at both model scales.}
At 4B, it reduces Average ASR from 6.65 to 3.84, a 2.81-point improvement over
the unadapted checkpoint, whereas LoRA reaches 5.33. The Safety State improves
four of the five jailbreak benchmarks, with only a slight 0.32-point regression
on StrongREJECT. At 35B-A3B, it improves all five benchmarks and lowers Average
ASR from 5.89 to 2.81, compared with 3.04 for LoRA. The largest per-benchmark
gain occurs on FORTRESS at both scales, with ASR reductions of 9.60 and 6.00
points at 4B and 35B-A3B, respectively.

\textit{(ii) The Safety State achieves a better safety--over-refusal
trade-off than the training-matched LoRA control.}
On XSTest, the Safety State increases ORR by 0.40 points at 4B (from 8.60 to
9.00) and by 2.00 points at 35B-A3B (from 17.60 to 19.60). Under matched
training conditions, LoRA produces substantially larger increases of 5.00 and
10.00 points, reaching 13.60 and 27.60, respectively. At the same time, the
Safety State achieves lower Average ASR than LoRA by 1.49 points at 4B and
0.23 points at 35B-A3B. These results show that the Safety State improves
jailbreak robustness while introducing substantially less over-refusal than
the LoRA control.

\textit{(iii) The Safety State preserves capabilities more consistently than
LoRA at 4B, while retention at 35B-A3B is task-dependent.}
At 4B, the Safety State remains within 2.60 points of the unadapted checkpoint
across all four capability benchmarks and outperforms LoRA on each of them.
At 35B-A3B, the results are mixed. The Safety State nearly matches the
unadapted checkpoint on MATH (93.32 versus 93.58) and retains more capability
than LoRA on MATH and AIME 2025, whereas LoRA performs better on MMLU-Pro and
MBPP. Thus, the Safety State provides a clear capability-retention advantage
over LoRA at 4B, while the larger model reveals no uniform winner across
capability tasks. Improving knowledge and code retention at the 35B-A3B scale
remains an important direction for future work.

Overall, Safety State provides a stronger safety--utility trade-off than matched LoRA.

\FloatBarrier

\section{Related Works}
\label{sec:related-work}

\paragraph{Efficient long-context sequence modeling.}
Efficient attention methods reduce the cost of full self-attention through
restricted receptive fields, kernel approximations, systems-level
optimization, or sparse selection. Local and sliding-window attention bound
the neighborhood available to each query
~\citep{wang2025rattentionminimalslidingwindow,
cabannes2025shortwindowattentionenables}, while Performer,
Nystr\"omformer, and Linear Attention replace or approximate the softmax kernel
with feature maps that permit associative computation
~\citep{performer,xiong2021nystromformer,katharopoulos2020transformers}.
FlashAttention-2 accelerates exact dense attention through hardware-aware
execution, while sequence-parallel algorithms improve the scaling of linear
attention~\citep{flashattention2,Sun2024LinearAS}. During autoregressive
decoding, however, exact dense attention still retains a token-level
key--value cache that grows with the context.

Sparse attention preserves content-based retrieval while reducing the number
of token-level interactions. Early methods impose predefined connectivity:
Sparse Transformer factorizes the attention graph, whereas Longformer and
BigBird combine local windows with global or random connections
~\citep{child2019generating,beltagy2020longformer,zaheer2020big}. Later systems
make selection input dependent. Routing Transformer clusters tokens by
content, H$_2$O evicts low-utility cache entries, and Quest selects KV-cache
pages according to the current query
~\citep{roy2021efficient,zhang2023h2o,tang2024quest}. Recent trainable designs
make routing part of the model itself: MoBA assigns queries to relevant
key--value blocks, while Native Sparse Attention combines local, compressed,
and selectively retrieved branches with hardware-aligned kernels
~\citep{lu2025moba,yuan2025native}. These methods differ in their selection
rules and retrieval granularity, but continue to operate over token- or
block-level key--value representations. \safinname{} instead retrieves compact
snapshots of the model's recurrent-state trajectory, placing long-range
routing over state-level memories.

\paragraph{Recurrent state and expandable memory.}
State-space models and gated linear recurrences compress the causal prefix
into a recurrent state, replacing an expanding token cache with fixed-size
updates and reads. Linear Attention exposes this connection through
outer-product state updates and query-based retrieval
~\citep{katharopoulos2020transformers,chou2024metala,dao2024transformers}.
Structured and selective state-space models improve the underlying transition
dynamics, while RetNet, RWKV, HGRN2, and related recurrent formulations combine
associative memory with structured recurrence
~\citep{gu2021efficiently,gu2023mamba,dao2024transformers,sun2023retentive,
peng2023rwkv,qin2024hgrn2,orvieto2023resurrecting,longhorn}. Gated Linear
Attention introduces input-dependent decay, and DeltaNet-style models use
delta-rule corrections to refine associative updates
~\citep{yang2024parallelizing,yang2025gateddelta,
hatamizadeh2026gateddeltanet2,siems2025deltaproduct,
Grazzi2024UnlockingSI}. Despite these advances, most recurrent models expose a
single fixed-capacity state, creating a retrieval bottleneck when earlier
associations are attenuated or overwritten
~\citep{arora2024zoology,arora2024simple,wen_rnns_2024}.

Expandable-memory methods address this bottleneck by increasing the number or
organization of accessible memory units. Multi-State RNNs, HGRN2, and
Log-Linear Attention maintain multiple or hierarchically organized recurrent
states~\citep{Oren2024TransformersAM,qin2024hgrn2,guo2026loglinear}. Other
systems decouple memory capacity from dense computation through routed memory
experts, sparsely accessed state banks, or product-key memories
~\citep{du2025mom,pan25SSE,lample2019largememorylayersproduct,
berges2024memorylayersscale,zhao2026fastweightproductkeymemory,cabannes26SDM,
afzal2026raven}. Context-compression methods use a different retrieval unit:
learned summary or gist tokens compress earlier spans, and selective schemes
reopen relevant compressed chunks when needed
~\citep{chevalier2023adapting,mu2023learning,zhang2024long,
deng2025unigist,petrov2025long,mao2026gisttokens}. These approaches therefore
span recurrent-state expansion, parameterized memory, and token-level context
compression.
MARCH~\citep{zhang2026march}, the architecture underlying \safinname{}, takes
a complementary approach: it preserves cumulative snapshots of the model's
native recurrent state, expands the
addressable state bank with context length, and retrieves historical states
through content-conditioned routing. The resulting memory units remain native
to the recurrent computation while becoming individually addressable by later
tokens. Unlike approaches that replace the recurrent state with a larger
learned memory, this design preserves the original update path while adding
addressability over its trajectory. State evolution and historical access are
therefore separated: the recurrence determines how the current state is
updated, while routing determines which earlier internal representations
become available to later tokens.

\Needspace{8\baselineskip}
\paragraph{Data-efficient and modular safety alignment.}
Safety alignment can be data efficient: Safety-Tuned LLaMAs show substantial
gains from a few hundred demonstrations, while STAR-1 aligns reasoning models
with 1K curated examples~\citep{bianchi2024safetytuned,wang2026star1}.
Robustness must nevertheless be evaluated alongside over-refusal, as
emphasized by adversarial suites such as WildJailbreak and benign-prompt
diagnostics such as XSTest
~\citep{jiang2024wildteaming,rottger2024xstest}. Parameter-efficient methods
preserve or restore safety through low-rank adaptation, pruning, or post-hoc
subspace composition~\citep{hsu2024safelora,li2025salora,ao2025splora,
zhou2025lssf,thakkar2025mergealign}. Together, these studies show that reducing
attack success is insufficient if alignment induces blanket refusal or erodes
general capability. They motivate our mixed harmful and benign supervision,
as well as the joint reporting of jailbreak success, over-refusal, and
capability retention.

A complementary line of work changes the locus of control from backbone
weights to model representations. ReFT learns
hidden-state interventions, while Representation Engineering, Contrastive
Activation Addition, and refusal-direction analyses manipulate behavior
through low-dimensional activation directions
~\citep{wu2024reft,zou2023representation,rimsky2024steering,
arditi2024refusal}. Safety-focused approaches transfer, sparsely adjust, or
input-condition these directions
~\citep{wang2024inferaligner,shen2025jailbreakantidote,lee2025cast,
zhao2025adasteer,wu2025safeint,jiang2026els}. Related modular work explores
other control interfaces: Circuit Breakers reroute internal representations,
MoGU routes among model variants, SAFEx studies safety-critical experts, and
Single Token Alignment uses a learned input prefix
~\citep{zou2024circuitbreakers,du2024mogu,lai2025safex,
yu2026singletoken}. \safinname{} differs primarily in the locus of
specialization. Its Safety State is learned from external harmful and benign
supervision, but is represented as a persistent matrix-valued recurrent state
and invoked through the model's native routing pathway alongside
context-derived memories. Unlike low-rank adaptation, it does not modify the
shared backbone; unlike per-token activation steering, it persists as a
routable memory candidate. The backbone remains frozen during specialization;
once learned, the Safety State can be attached or removed, and the router
assigns it a token- and context-dependent contribution. This provides a
state-native interface for test-time adaptation.

\FloatBarrier

\section{Discussion and Conclusion}
\label{sec:conclusion}

\paragraph{Model memory as an architectural substrate.}
\safinname{} demonstrates that recurrent state can serve not only as a
transient summary of the prefix, but also as an addressable substrate for
long-horizon memory. Its underlying MARCH state-routing mechanism preserves
selected points along the recurrent trajectory and makes them individually
addressable. The consistent gains across GDN, KDA, and GDN2 indicate that
historical-state access is complementary to the recurrent update rule, while
the 4B and 35B-A3B \safinname{} experiments show that these benefits persist
after continual pretraining and supervised fine-tuning. Recurrent memory
quality therefore depends not only on how information is written into the
current state, but also on whether useful earlier versions remain accessible.
Because state anchors are compressed cumulative memories rather than
token-level records, this design complements rather than replicates the exact
recall path of full attention.

\paragraph{State-native specialization for Safety from Within.}
In \safinname{}, the same routed state bank can host both context-derived
anchors and learned persistent capability states. Safety State provides an
initial demonstration: with the backbone frozen, it is retrieved through the
native router and achieves a better aggregate safety--over-refusal trade-off
than training-matched LoRA. Here, \emph{Safety from Within} refers to where
safety is represented and invoked, not to where its supervision originates:
the state is still learned from external harmful and benign examples. Its
ability to be attached or removed enables modular specialization over a shared
backbone, but does not make safety immutable or universally robust.
Detachability also creates an integrity requirement: both the persistent state
and its routing path must be protected against removal, substitution, or
adversarial suppression. The same interface may extend to capabilities beyond
safety, although whether multiple persistent states can coexist and compose
reliably remains an open question.

\paragraph{Limitations and research agenda.}
Architecturally, the state bank grows with context length, and sparse routing
reduces reading cost but not state storage; fixed-interval anchoring may also
retain redundant checkpoints. Empirically, our large-scale evidence is limited
to two Qwen3.5 configurations under a single CPT--SFT pipeline. The safety
study uses a small corpus and a limited set of English benchmarks, does not
cover adaptive or multilingual attacks or state tampering, and shows
task-dependent capability retention at 35B-A3B. Future work should develop
adaptive anchor creation, consolidation, and eviction; evaluate broader model
families and context regimes; and study how multiple persistent states can be
composed without routing conflicts or capability interference. Beyond
improving the current state-routing design, it is essential to investigate
alternative and complementary architectures that make safety representable,
maintainable, and selectively invocable through a model's native computation.

\paragraph{Conclusion.}
We introduced \safinname{}, a family of foundation models that treats model
state as both addressable context memory and an interface for adaptable
capabilities. Controlled 0.8B experiments validate the underlying
state-routing mechanism across recurrent update rules, while matched 4B and
35B-A3B CPT--SFT experiments show that its gains persist at larger scale. With
the backbone frozen, Safety State reduces the unweighted average ASR across
five jailbreak benchmarks by 42.3\% and 52.3\% at 4B and 35B-A3B,
respectively, while yielding lower XSTest ORR than the training-matched rank-8
LoRA controls. These results provide initial architecture-level evidence for
\emph{Safety from Within}, rather than establishing a definitive solution.
Exploring alternative and complementary architectural routes toward this
broader objective remains essential.
\par


\makeatletter
\fancypagestyle{authorshipstyle}{%
    \fancyhf{}%

    \fancyhead[L]{\GDM@maybeLeftLogo}%
    \fancyhead[C]{}%
    \fancyhead[R]{\GDM@maybeRightLogo}%

    \fancyfoot[L]{}%

    \fancyfoot[C]{}%

    \fancyfoot[R]{%
        \footerfont\thepage
    }%
}
\makeatother

\FloatBarrier
\section*{Project Contributors}
\phantomsection
\addcontentsline{toc}{section}{Project Contributors}

This work was led by the Shanghai Artificial Intelligence Laboratory,
with contributions from the authors listed below in contribution order.

\begin{center}
    \renewcommand{\arraystretch}{1.35}
    \setlength{\tabcolsep}{6pt}
    \begin{tabularx}{\textwidth}{
        @{}
        *{4}{>{\centering\arraybackslash}X}
        @{}
    }
        Ming Zhang$^{*}$
        & Kaisen Yang$^{*}$
        & Shu Yu
        & Ermo Hua
        \\

        Zhekai Chen
        & Cheng Jin
        & Jingnan Zheng
        & Yi Zhang
        \\

        Dongcheng Zhang
        & Zhongtian Ma
        & Jiawei Zhou
        & Sirui Chen
        \\

        Qiaosheng Zhang
        & Xiang Wang
        & Ning Ding
        & Xia Hu
        \\

        Bowen Zhou
        & Youbang Sun$^{\ddagger}$
        & Chaochao Lu$^{\dagger}$
        &
    \end{tabularx}

    \vspace{0.5em}
    \small
    \textbf{$^*$ Equal Contribution.}~~
    \textbf{$^\ddagger$ Technical Lead.}~~
    \textbf{$^\dagger$ Project Lead.}
\end{center}

\thispagestyle{authorshipstyle}

\FloatBarrier
\Needspace{12\baselineskip}
\begingroup
\sloppy
\phantomsection
\printbibliography[heading=bibintoc]
\endgroup

\appendix

\section{Training Configurations}
\label{app:training-configurations}
\pagestyle{fancy}

This appendix reports the architecture, data, and optimization configurations
for the controlled small-scale study, the 4B and 35B-A3B \safinname{} runs,
and Safety State specialization.

\subsection{Small-Scale MARCH Validation}
\label{app:small-scale-training}

\begin{table}[H]
  \centering
  \caption{Training configuration for the controlled small-scale study.}
  \label{tab:small-scale-training-config}
  \small
  \renewcommand{\arraystretch}{1.12}
  \setlength{\tabcolsep}{6pt}
  \begin{tabularx}{\textwidth}{@{}>{\raggedright\arraybackslash}p{0.29\textwidth}X@{}}
    \toprule
    \textbf{Configuration} & \textbf{Setting} \\
    \midrule
    Training regime & From scratch \\
    Training corpus
      & \href{https://huggingface.co/datasets/togethercomputer/Long-Data-Collections}{Long-Data-Collections} \\
    Token budget & 50B tokens \\
    Sequence length & 16,384 tokens \\
    Global batch size
      & 256 packed sequences (4.19M token slots per update) \\
    Optimizer
      & Fused AdamW ($\beta_1=0.9$, $\beta_2=0.95$,
        $\epsilon=10^{-8}$) \\
    Peak / minimum learning rate
      & $4\times10^{-4}$ / $4\times10^{-5}$ \\
    Learning-rate schedule
      & WSD: 256-step warmup, 10,473 stable steps, and 1,192 decay steps \\
    Weight decay & 0.1 \\
    Gradient clipping & Global $\ell_2$ norm of 1.0 \\
    Precision & BF16 model computation; FP32 optimizer states \\
    \bottomrule
  \end{tabularx}
\end{table}

\Needspace{7\baselineskip}
\paragraph{Shared architecture.}
The recurrent models in the primary controlled comparison use a 21-layer
fully recurrent backbone with hidden size 1,536, a SwiGLU intermediate size
of 4,096, and untied input and output embeddings. All models employ the
Llama 2 tokenizer~\citep{touvron2023llama2} with a vocabulary size of 32,000.
GDN, KDA, and GDN2 use six recurrent heads with $d_k=192$ and $d_v=384$,
together with a bias-free short convolution of kernel size 4.

For \sysname{}, state anchors are created every $C=512$ tokens and all
causally visible anchors participate in dense routing. The router uses six
query heads, one key head, routing dimension $d_r=64$, and a learned null
candidate. The Log-Linear baseline retains at most 15 hierarchical states.

\FloatBarrier
\subsection{Large-Scale \safinname{} Training}
\label{app:safin-training}

\paragraph{Backbones and state-routing configuration.}
We initialize the dense 4B model from Qwen3.5-4B and the MoE model from
Qwen3.5-35B-A3B-Instruct. Both backbones use a 3:1 ratio of GDN to
full-attention layers. At both scales, 16 selected GDN layers within layers
10--30 (one-based) are equipped with state anchoring and routed historical
readout, with anchors created every $C=512$ text tokens. The router uses 32
query and 32 key heads with $d_r=128$ per head, selects the Top-$4$ candidates,
and includes a learned null candidate. Historical readout reuses the native
GDN query, without a separate historical-read query projection or route
positional encoding.

\paragraph{Initialization.}
The anchor embedding is initialized as $0.5\,e_{\mathrm{BOS}}$. Anchor-query
and anchor-key projections are initialized from the corresponding pretrained
GDN projections after head expansion, and the routing-query projection is
copied from the expanded anchor-query projection. Routing RMSNorm parameters
are copied from the input layer normalization, while the null projection and
bias are initialized to zero. All remaining backbone weights are loaded from
the starting checkpoint.

\paragraph{Training protocol.}
Within each scale and training stage, Qwen3.5 and \safinname{} use the same
training data, token budget, and optimization settings. All runs use Megatron
distributed Adam with $\beta_1=0.9$, $\beta_2=0.95$, $\epsilon=10^{-8}$,
weight decay 0.1, and global gradient clipping at 1.0. Model parameters and
forward--backward computation use BF16, with FP32 optimizer states and
numerically sensitive operations; the 35B-A3B MoE router also remains in FP32.
Global token batches count text-token slots and exclude inserted anchor tokens.
During SFT, all language-model parameters are updated while the vision tower
and aligner remain frozen, and state-routing parameters are inherited from CPT
without reinitialization.

Tables~\ref{tab:large-scale-model-config} and
\ref{tab:large-scale-training-config} summarize the scale-specific model and
training configurations.

\begin{table}[!htbp]
  \centering
  \caption{Model configurations for the 4B and 35B-A3B \safinname{} models.}
  \label{tab:large-scale-model-config}
  \scriptsize
  \renewcommand{\arraystretch}{1.0}
  \setlength{\tabcolsep}{4pt}
  \begin{tabularx}{\textwidth}{@{}>{\raggedright\arraybackslash}p{0.27\textwidth}
      >{\centering\arraybackslash}X>{\centering\arraybackslash}X@{}}
    \toprule
    \rowcolor{gray!12}
    \textbf{Configuration} & \textbf{4B} & \textbf{35B-A3B} \\
    \midrule
    CPT initialization
      & Qwen3.5-4B
      & Qwen3.5-35B-A3B-Instruct \\
    Decoder layout
      & 32 layers (24 GDN + 8 full-attention)
      & 40 layers (30 GDN + 10 full-attention) \\
    Hidden size
      & 2,560 & 2,048 \\
    GDN state configuration
      & \multicolumn{2}{c}{32 value heads; 16 key heads; $d_k=d_v=128$} \\
    State-routing layers
      & \multicolumn{2}{c}{16 selected GDN layers within layers 10--30 (one-based)} \\
    \bottomrule
  \end{tabularx}
\end{table}

\begin{table}[H]
  \centering
  \caption{Training configurations for continual pretraining and supervised
  fine-tuning.}
  \label{tab:large-scale-training-config}
  \scriptsize
  \renewcommand{\arraystretch}{0.98}
  \setlength{\tabcolsep}{4pt}
  \begin{tabularx}{\textwidth}{@{}>{\raggedright\arraybackslash}p{0.31\textwidth}*{2}{>{\centering\arraybackslash}X}@{}}
    \toprule
    \rowcolor{gray!12}[0pt][0pt]
    \multicolumn{3}{@{}l@{}}{\textbf{Continual Pretraining}} \\
    \textbf{Configuration} & \textbf{4B} & \textbf{35B-A3B} \\
    \midrule
    Training data
      & \multicolumn{2}{c}{Intern-S2 pretraining mixture~\citep{bai2026interns2preview}} \\
    Training tokens
      & \multicolumn{2}{c}{50.004B} \\
    Maximum sequence length
      & \multicolumn{2}{c}{32,768 text tokens} \\
    Training steps
      & \multicolumn{2}{c}{5,961} \\
    Global batch
      & \multicolumn{2}{c}{256 sequences (8.389M token slots)} \\
    Peak / minimum learning rate
      & \multicolumn{2}{c}{$5\times10^{-5}$ / $10^{-6}$} \\
    Learning-rate schedule
      & \multicolumn{2}{c}{WSD: 256-step warmup, 5,109 stable steps, and 596 decay steps} \\

    \midrule
    \rowcolor{gray!12}[0pt][0pt]
    \multicolumn{3}{@{}l@{}}{\textbf{Supervised Fine-Tuning}} \\
    \textbf{Configuration} & \textbf{4B} & \textbf{35B-A3B} \\
    \midrule
    Training data
      & \multicolumn{2}{c}{\href{https://huggingface.co/datasets/nvidia/Nemotron-Cascade-2-SFT-Data}{Nemotron-Cascade-2-SFT-Data}~\citep{yang2026nemotroncascade2}} \\
    Training tokens
      & \multicolumn{2}{c}{30.199B} \\
    Maximum sequence length
      & \multicolumn{2}{c}{32,768 text tokens} \\
    Training steps
      & 3,600 & 7,200 \\
    Global batch
      & 256 sequences (8.389M token slots)
      & 128 sequences (4.194M token slots) \\
    Peak / minimum learning rate
      & \multicolumn{2}{c}{$10^{-5}$ / $10^{-6}$} \\
    Learning-rate schedule
      & 5\% warmup (180 steps), then cosine decay
      & 5\% warmup (360 steps), then cosine decay \\
    \bottomrule
  \end{tabularx}
\end{table}

\FloatBarrier

\subsection{Safety State Specialization}
\label{app:safety-state-training}

We compare the persistent Safety State with a rank-8 LoRA control under a
matched adaptation protocol. Both methods start from the corresponding
\safinname{} SFT checkpoint and use the same examples, optimization schedule,
batch size, and 37 updates. For Safety State, we freeze the backbone and
optimize only the persistent states. For LoRA, we disable Safety State and
update only the adapters while retaining dynamic state anchoring and
content-routed historical readout.

\begin{table}[H]
  \centering
  \caption{Shared adaptation configuration for Safety State and the rank-8
  LoRA control.}
  \label{tab:safety-state-training-config}
  \footnotesize
  \renewcommand{\arraystretch}{1.08}
  \setlength{\tabcolsep}{5pt}

  \begin{tabularx}{\textwidth}{@{}>{\raggedright\arraybackslash}p{0.29\textwidth}X@{}}
    \toprule
    \textbf{Configuration} & \textbf{Setting} \\
    \midrule
    Training data
      & 1,000 STAR-1 harmful and 915 STAR-benign targets \\
    Context conditions
      & 0-, 512-, and 2,048-token benign prefixes; 5,745 examples in total \\
    Starting checkpoint
      & Corresponding final \safinname{} SFT checkpoint \\
    Training budget
      & 1 epoch; 37 optimizer steps; global batch of 32 packed sequences \\
    Optimizer
      & Megatron distributed Adam ($\beta_1=0.9$, $\beta_2=0.95$,
        $\epsilon=10^{-8}$) \\
    Learning-rate schedule
      & Peak $3\times10^{-4}$; minimum $3\times10^{-5}$; 10\% warmup followed by cosine decay \\
    Regularization
      & Weight decay 0.01; global gradient clipping 1.0 \\
    Maximum sequence length
      & 8,192 tokens \\
    Numerical precision
      & BF16 model computation and LoRA adapters; FP32 Safety State parameters
        and optimizer states \\
    \bottomrule
  \end{tabularx}
\end{table}

\FloatBarrier

\FloatBarrier

\end{document}